# The State of Industrial Robotics: Emerging Technologies, Challenges, and Key Research Directions

**now**

the essence of knowledge

Boston — Delft

# Contents

# The State of Industrial Robotics: Emerging Technologies, Challenges, and Key Research Directions

Lindsay Sanneman[*1], Christopher Fourie[*2] and Julie A. Shah[3]

[1]*Computer Science and Artificial Intelligence Laboratory (CSAIL), Massachusetts Institute of Technology (MIT), 77 Massachusetts Ave, Cambridge, MA 02139, USA; lindsays@csail.mit.edu*
[2]*Computer Science and Artificial Intelligence Laboratory (CSAIL), Massachusetts Institute of Technology (MIT), 77 Massachusetts Ave, Cambridge, MA 02139, USA; ckfourie@csail.mit.edu*
[3]*Computer Science and Artificial Intelligence Laboratory (CSAIL), Massachusetts Institute of Technology (MIT), 77 Massachusetts Ave, Cambridge, MA 02139, USA; julie_a_shah@csail.mit.edu*

ABSTRACT

Robotics and related technologies are central to the ongoing digitization and advancement of manufacturing. In recent years, a variety of strategic initiatives around the world including "Industry 4.0", introduced in Germany in 2011 have aimed to improve and connect manufacturing technologies in order to optimize production processes. In this work, we study the changing technological landscape of robotics and "internet-of-things" (IoT)-based connective technologies over the last 7-10 years in the wake of Industry 4.0. We interviewed key players within the European robotics ecosystem, including robotics manufacturers and integrators, original equipment manufacturers (OEMs), and applied industrial research institutions and synthesize our findings in this paper. We first detail the state-of-the-art robotics and IoT technologies we observed and that the companies discussed during our interviews. We then describe the processes the companies follow when deciding whether and how to integrate new technologies, the challenges they face when integrating these technologies, and some immediate future technological avenues they are exploring in robotics and IoT. Finally, based on our findings, we highlight key research directions for the robotics community that can enable improved capabilities in the context of manufacturing.

---
[*]These authors contributed equally to this work.

# 1

## Introduction

Over the last decade, a variety of initiatives and frameworks for advancement of manufacturing technologies have been proposed around the world, including in the United States, India, China, Russia, and several countries within the European Union (EU) (Henning, 2013). One early example of such a framework is Industry 4.0, which was introduced as a strategic initiative in Germany during 2011 and has since been adopted internationally (Lobova *et al.*, 2019; Lu, 2017). Industry 4.0, also known as the "Fourth Industrial Revolution," focuses primarily on advancement of cyber-physical systems (CPS), which Rajkumar *et al.* (2010) defined as "physical and engineered systems whose operations are monitored, coordinated, controlled and integrated by a computing and communication core." It has emerged as both a reaction to the digitization of manufacturing and an economic and technological driver inspiring the creation of technologies for the digitization of manufacturing.

According to Lu (2017), the First Industrial Revolution involved water- and steam-powered mechanical production plants at the end of the 18th century, the Second Industrial Revolution involved a transition to electrically-powered mass production, and the Third Industrial Revolution leveraged electronics and information technology to automate production. The aim of Industry 4.0 is to improve and connect automated systems through advancement of CPS in order to achieve a higher level of operational efficiency, productivity, and interconnectedness within factories, enhancing the optimization of production processes (Lu, 2017).

Central to the vision of Industry 4.0 in the manufacturing context are both manufacturing technologies (such as robotics and other automated systems) and the platforms, infrastructures, and systems that enable coordinated control and connection between them, collectively deemed the "internet of things" (IoT). In this work, we study the technological development of robotic technologies (including both industrial and service robots) and IoT-based technologies (such as cloud computing systems, simulation systems, and data infrastructures) and their adoption within the context of Industry 4.0. In particular, we explore whether such technologies have manifested tangible differences from past capabilities, and whether these technologies have now begun to or have the potential to provide their predicted benefits, 7-8 years after the introduction of Industry 4.0. We study the European context, with a heavy focus on Germany, because the term "Industry 4.0" originated there and has the highest relative presence in normative and legal state documents in Germany compared with other countries that also adopted the term (Lobova *et al.*, 2019), implying that governmental and legislative support for the strategy has been strong there. Further, compared with its European counterparts, Germany has the highest density of industrial robots at 309 per 10,000 employees (International Federation of Robotics, 2016).

We interviewed key players within the industrial robotics ecosystem in Germany and nearby European countries in order to investigate the extent to which the technological landscape has changed since the introduction of Industry 4.0. Our interview subjects included representatives from robotics manufacturers and integrators (companies with a dedicated interest in the development and integration of industrial robots); original equipment manufacturers (OEMs), which leverage industrial robots for automated assembly lines; and industrial research institutions which focus on general research challenges related to industrial automation and work with OEMs on domain-specific automation solutions. Our questions pertained to company details and ecosystem relationships, emerging technologies, IoT-style integration and Industry 4.0, standardization of technologies, the human line worker's role with regard to new technologies, metrics for robotic solutions, potential next directions for technological development, and the challenges companies face related to the integration of new technologies.

Previous works have included interviews of manufacturing employees working with new robotics technologies to better understand the adoption of these technologies in real-world scenarios, with a primary focus on line workers and those closely interacting with the technology on a daily basis (Elprama *et al.*, 2016; Elprama *et al.*, 2017; Sauppé and Mutlu, 2015; Welfare *et al.*, 2019; Wurhofer *et al.*, 2018). While we consider the line-worker perspective to be valuable and important, we also note the merits of understanding broader ecosystem drivers of new technologies as they relate to the evolving human role in manufacturing. In our approach, we focus on which technologies have been successfully demonstrated and which factors limit the application of new technologies within the industrial robotics ecosystem. We also highlight issues around the integration of these technologies, manufacturer expectations, and challenges around standardization and present key directions for future research as highlighted by industry representatives and stemming from current problems in the application of these technologies.

The remainder of the paper is structured as follows: in Section 2, we detail our interview process, including the conducting and analysis of interviews. In Section 3, we discuss current and emerging technologies addressed in the interviews or demonstrated during post-interview tours of manufacturing or research facilities. Section 4 describes the processes the interviewed companies follow while deciding when and how to introduce new technologies. Section 5 details some of the challenges the companies face when implementing new robotic solutions. Section 6 lists the primary future directions the companies discussed during interviews. In Section 7, we enumerate some potential directions for robotics research based on the synthesis of key ideas gleaned from the interviews. Finally, Section 8 discusses work related to the study performed in this work, and Section 9 concludes the paper.

# 2

## Study Process

We interviewed senior management/engineering staff and/or research directors at four leading robotics manufacturers and integrators (including one emerging robotics developer), two applied research institutes, and three large multinational manufacturers (one in the automotive industry). We conducted the interviews at on-site locations in Germany, France, and Italy. These semi-structured interviews were guided by a set of seven questions related to established technologies, standardization, metrics, the changing nature of human work, and immediate directions for future technology development. The interviews typically lasted 1-2 hours, and most included additional tours and demonstrations of the technology before or after the interview.

We recorded detailed notes during the interviews, and followed a qualitative research approach known as "thematic analysis", wherein interviews are organized according to key themes, to extract information related to our particular areas of interest while preserving context. We used qualitative research software (`Quirkos` (Turner, 2020)) to organize our notes, assigned codes to the text according to themes that emerged during our research, and then synthesized the information into the topics presented here.

The approach has some limitations. First, interpretation is subjective and the information is qualitative; as such, the reader is advised against interpreting the frequency at which information appears as an indication of its importance or relevance. Also, while the authors possess a background in robotics, we also recognize that subjectivity in interpretation may influence our results. Finally, due to the relatively small number of large robot developers, manufacturers with a high number of robots on their factory floor, and industrial research institutions focused on robotics, our samples are not random; this is an evaluation based on expert opinions.

## 2.1 Company Selection Process

We selected companies for participation based on their involvement with industrial robotics in continental Europe, particularly Germany, France, and Italy. The robot manufacturers range from small, emerging developers to large, well-known members of the ecosystem. Product manufacturers and applied research institutes that employ robots were chosen based on investment into industrial robotics. We scheduled interviews either through direct contact or via a network of academic contact points.

## 2.2 Companies Interviewed

We performed nine interviews in total; the relevant companies are numbered and briefly described below. Robot manufacturers are identified with "RM," applied research institutions with "RI," and product manufacturers with "PM."

- [RM1] - A large company focused on the manufacturing of robotics technologies, ranging from light- to heavy-payload robotic arms, as well as various service robots and collaborative systems. The company is also a well-known integrator that provides automation services internationally.

- [RM2] - A large company focused on manufacturing robotics technologies ranging from light- to heavy-payload robotic arms, as well as industrial research into collaborative systems. The company is also a well-known integrator, providing automation services internationally.

- [RM3] - A medium-sized company that manufactures small- to medium-payload industrial robots. The company also provides tailored surface coatings and finishings for robots to support a variety of specialized applications.

- [RM4] - A small company focused on the manufacturing and development of lightweight-compliant robotic arms, as well as innovative cloud technology and robot teaching methods.

- [RI1] - A large applied research institution dedicated to automation of manufacturing. The institute has significant experience developing state-of-the-art technologies for industrial and service robots for direct industrial application.

- [RI2] - An applied research institution focused on machine tools and automation. The institute has significant experience in the development of collaborative robot systems for industrial application.

- [PM1] - A large, multinational manufacturer of a variety of consumer products, including power tools and large appliances.

- [PM2] - A large, multinational manufacturer of industrial electronics products, ranging from electronic controllers to electrical distribution.

- [PM3] - A large automotive manufacturer that produces a variety of luxury consumer vehicles.

## 2.3   Interview Process and Questions

All interviews were semi-structured and involved questions across seven categories: company details and ecosystem relationships, emerging technologies, IoT-style integration and Industry 4.0, standardization, the human role in the context of new technologies, metrics for robotic solutions, and technological next directions and challenges within the context of automation. Interviews lasted 1-2 hours apiece, and seven of the nine also included a tour of a manufacturing facility or research space. Interview subjects were promised confidentiality and the opportunity to review information reported here prior to publication. Audio was recorded for eight of the nine interviews.

The authors of this paper conducted all interviews and associated correspondence. The same researchers were present at all nine interviews, all of which included the same seven questions. While the same high-level question structure was followed in all cases, the interviewers asked follow-up questions when beneficial, and subjects were able to elaborate upon points they deemed important. The interviewers asked the following questions:

1. Tell us about your organization. Which manufacturing companies, integrators, robotics manufacturers, or applied research institutions do you work with (to the extent that you can share)? What do these collaborations look like?

2. What kinds of technologies were you developing in this division 7-10 years ago? What are you working on now?

   (a) What were the primary drivers of any shifts in direction (industry, government policies, company interests, etc.)?

   (b) What have you seen actually adopted in industry? Has this changed much? What allowed this (or if not, what challenges prevented this)?

   (c) Have there been measurable changes in metrics like productivity, efficiency, quality, etc. as a result of new technologies?

3. Are technologies developed today more integrated with other systems than those developed 8-10 years ago (IoT-style integration)? What do you think is the reason for this change (if one exists)? What does this integration look like for you (if applicable)?

4. Do you see more of a shift toward standardization now than in the past (7-10 years ago)? What kinds of things are being standardized? Why do you think this is the case (if it is), or what do you think is inhibiting standardization (if it is not)?

5. Has the role of the person interfacing with the machinery changed in the last 7-10 years? Are you seeing an increased uptake in collaborative systems? Are there new styles of work? Has your number of workers changed?

6. What metrics are used to assess the performance/value of robotics? Do you see a change in the metrics used to describe the effect of robotics? Has the sales pitch changed?

(a) In the case of collaborative systems, are there new metrics companies use to assess value proposition?

7. What do you see as some of the immediate next directions in terms of new technologies? What do you perceive the greatest challenges in industrial and collaborative robotics to be?

# 3

# Emerging Technologies in Manufacturing

In this section, we define and discuss current and emerging technologies that that were observed on the factory floor and were discussed during the interviews, their stated uses, and the associated perceptions of them held by the interviewees. The technologies discussed do not represent a comprehensive or exhaustive list, but rather a representation of those observed during the interview process. We augment the discussion where necessary with definitions and descriptions drawn from available literature. This discussion is split into two categories: emerging technologies surrounding the use of industrial and service robots, and those around systems architectures (e.g., the Internet of things [IoT]) and general automation enhancements.

## 3.1 Industrial Robotics Technologies

In this work, we employ the distinction between industrial and service robots established by the International Federation of Robotics (IFR) (IFR, 2019) and the International Organization of Standardization (ISO): an industrial robot is "*an automatically controlled, reprogrammable multipurpose manipulator programmable in three or more axes,*" while a service robot is a robot "*that performs useful tasks for humans or equipment, excluding industrial automation applications*" (ISO, 2012).

Industrial robots must operate for long periods of time without failure. During interviews, [RM3] and [RM1] stated that minimum targets for operation are typically on the order of 16,000 hours, with some products able to last as long as 80,000 hours. [RM1] explained that robotic arms have a fairly long innovation cycle: approximately 7 years. Recent innovation in industrial robotics has primarily occurred in the context of improved safety systems or integrations with additional sensing mechanisms, as opposed to innovation of the robotic arm itself. We discuss these complementary systems in detail in the following subsections.

### 3.1.1   Safety Systems

An industrial robot can be dangerous to nearby humans, and safety systems seek to ensure that no physical or psychological harm can take place through inadvertent contact with a person operating in close proximity to a robot (Lasota *et al.*, 2017). Traditionally, a cage around a production cell containing a robot (as shown in Figure 3.1) ensures that a human cannot access that area while the robot is in operation - improving safety but increasing the cell's footprint. Emerging safety systems allow for reduction of a production cell's footprint via robotics technologies, in some cases removing the need for cages altogether. These technologies fall into two categories: passive safety systems (which do not actively detect the presence of humans) and active safety systems (which incorporate sensors that detect a human's proximity). Passive systems operate either through sensors that trigger a stop when a robot collides with a human or object within the environment, or through design properties deemed safe for human interaction (such as a limitation on the force the robot can apply). Active systems involve the integration of sensors that fall into two categories: standard industrial sensors (such as light curtains, as depicted in Figure 3.2, or proximity sensors within the cell) or active sensors on the robot itself (torque sensors and capacitive skins on the robotic arm). These safety systems typically require conformance to the ISO15066 safety standard (ISO, 2016), which limits a robot's overall velocity and the force with which a robot may interact with a person. As this can limit the speed at which a robot can operate in a manufacturing cell, the benefit of applying these technologies to create shared workspaces (collaborative systems) is debated.

During interviews, [PM1] stated that "collaborative robotics is slow-motion robotics," implying that reduced velocities can pose limitations in scenarios where they are not required, while [RM4] and [RM3] pointed out that current use cases are limited. However, [PM3] noted that many human-robot interaction applications exist, and [PM1] said the mindset regarding collaborative robots is changing, and that situations exist where a robot can act in order to assist a human worker.

Passive safety systems that are not payload restricted typically involve sensors that support collision detection. Passive robot skins, such as those constructed from polyimide (Duchaine *et al.*, 2009) or electrically insulative films (Alvite, 1987), detect when the skin encounters an obstacle, enabling the robot to trigger an emergency stop. [RM3] stated that use of this technology is emerging, but its relatively high expense limits demand.

Alternative approaches to safety lie in the design of the robotic arm. Smaller, payload-limited arms, such as Universal Robotics' UR10 Robotic Arm (Universal Robotics, 2020), Franka Emika's Panda Arm (Frank Emika, 2020), or KUKA's IIWA arm (Kuka, 2020) (collectively referred to as "lightweight robot arms") ensure safety through conformance to the ISO15066 standard (ISO, 2016) by limiting the possible force that a robot could apply to a human. Another example of passive safety systems is torque feedback, or *compliance*. Many lightweight robotic arms implement compliance, which can further limit the force experienced by the human in the event of a collision by allowing the arm to deflect when encountering an unexpected obstacle.

Active safety systems detect the presence of a human before a collision can occur. Capacitive robot skins, such as the one depicted in Figure 3.3, shroud the robot with a sensory film that identifies electrical field changes within a short range around the robot (10-20 cm to detect the presence of an obstacle (a human, in this context)) (Ulmen and Cutkosky, 2010). As stated by [PM1], this technology negates the need for torque-limited arms in a collaborative context; however, [RM3] noted this technology is not completely mature (for example, the skins have been known to cause the robotic arm to detect itself and lock in place).

Other active safety systems involve standard industrial sensors (such as a light curtain or laser-based proximity sensor), creating a cell where the speed of the robotic arm is dynamically modified based on its proximity to the detected individual. We observed examples of these systems on the production lines of some manufacturers featured in this work ([RM3] and [RM1]).

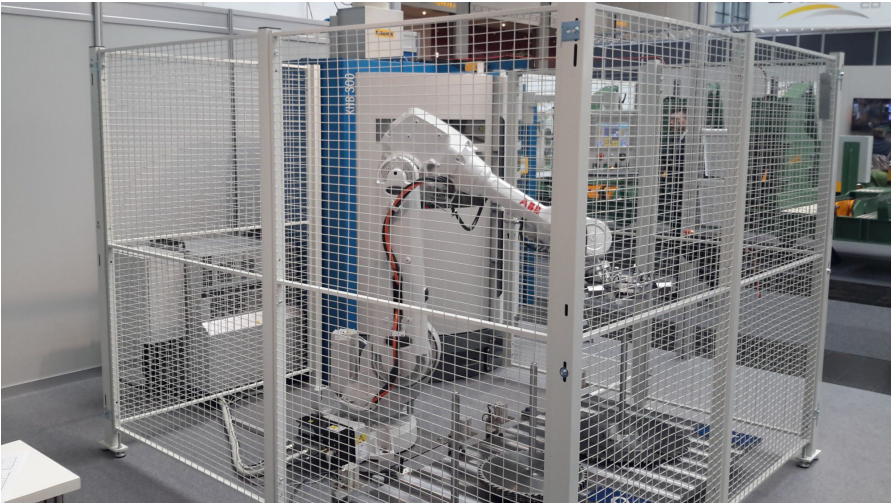

**Figure 3.1:** Industrial robot cell with a cage (Padget Technologies, 2020)

### 3.1.2   Collaborative and Lightweight Arms

An industrial, collaborative robotic system is one in which a human associate can safely enter a robot's work zone; however, this does not necessarily imply that the human and robot collaborate directly on a particular workpiece. We consider lightweight arms to be payload-limited robotic arms, rather than their integration into industrial collaborative systems (although many are used in such a manner, as the technology can ensure the safety of a human in close proximity to the robot).

With regard to collaborative systems, [PM1] notes that such systems are not always desirable and that cages can be beneficial as they stop a human from disrupting the robot's work. However, [PM1] also notes that the reduction in workcell footprint resulting from the integration of safe, collaborative systems can be beneficial; but, as outlined by both [RM1] and [PM2], it is not the robot that must be considered safe for the human to interact with or work around, but the full system. The application of collaborative systems is to tasks which are inherently safe, and, as many of the tasks that robots typically perform are dangerous (e.g. moving objects with sharp edges), this can limit the application of collaborative systems. The topic of direct collaboration appears to be emerging: few collaborative systems were observed, and those that were involved the robot slowing down or stopping as a human associate collected completed parts from the workcell.

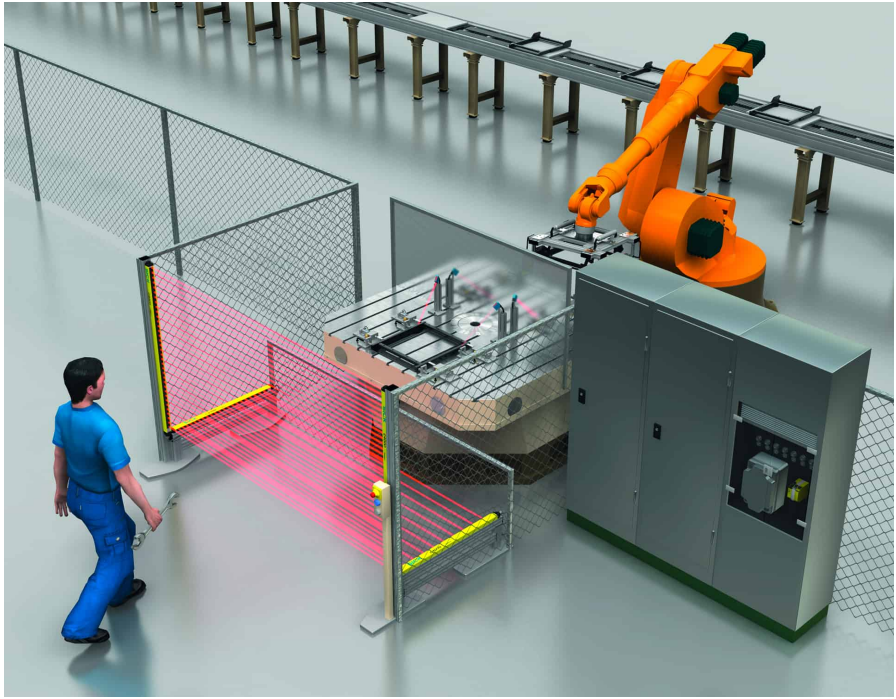

**Figure 3.2:** Depiction of a light curtain safety system (SICK (UK) LTD, 2017)

Lightweight robotics are focused on technologies with a reduced payload. These robots are low-cost, and are believed by several interviewees to be breaking into new sectors of the market. [PM1] and [PM2] highlighted lightweight robotics as facilitating the "democratization of robotics." The applications of these machines remain simple, largely due to integration costs; however, [PM2] noted that the relative safety of these devices enabled simpler testing and integration, facilitating their integration into domains where standard industrial arms are too dangerous for engineers to directly work with. [RI1] and [RM3] said that lightweight robot arms can still be hazardous despite payload limitations, and [RM3] suggested that their relative success lies in their ease of programming and low cost.

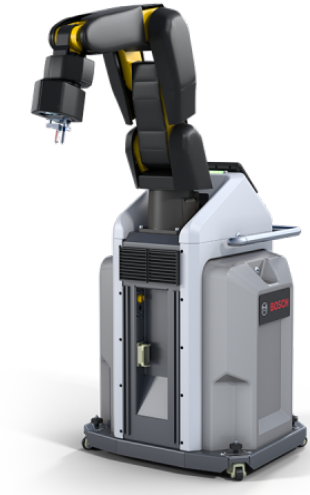

**Figure 3.3:** Robot shrouded in a capacative skin (Universita Degli Studi Di Brescia, 2020)

### 3.1.3   Compliant Systems

Compliant arms, such as the one shown in Figure 3.4, utilize integrated torque sensors for improved interaction capabilities (De Schutter and Van Brussel, 1988). While this technology has existed since the late 1980's, recent developments appear to be improving their uptake. Both [PM3] and [RM1] reported using these devices to perform complex manipulation involving loose-fitting parts, while [PM3] used compliant technologies in some collaborative systems. Compliant capabilities promise to not only improve safety by ensuring obstacles are detected and the robot's motion is constrained, but also to enable new capabilities that were not possible before (e.g., insertion). [PM1] noted that errors during sensitive manipulations represent limitations in manufacturing, and that compliant robotics overcome these limitations.

However, demand for this technology is still emerging: [RM3] claimed to possess the technology to build such machines but could not yet justify the investment, while some tasks remain out of reach despite the use of compliant robots (for example, [PM1] reported difficulty implementing systems to manipulate screws). [RM4], in contrast, considered compliance a core focus of their technology development, while [PM1] cited it as a key enabling technology in the future.

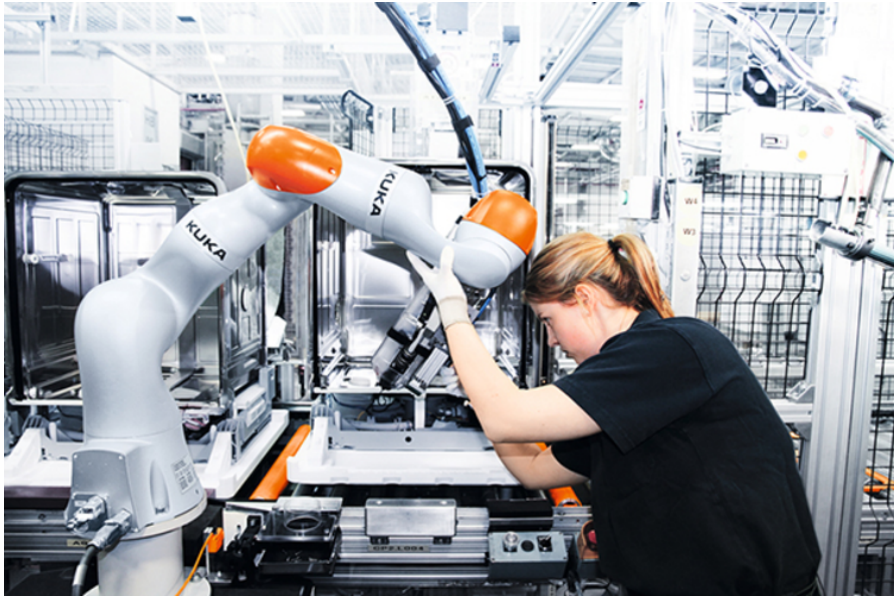

**Figure 3.4:** Collaborative robot arm implementing compliance (Robotics Business Review, 2018)

### 3.1.4 Gripping

Robotic gripping (for example, see Figure 3.5) involves the physical manipulation and grasping of workpieces in a factory workcell. As stated by [PM1], robotic gripping capabilities are far from those of humans, and physical gripping hardware remains an enormous challenge. Generalized bin picking, wherein parts are not precisely ordered and arranged, was cited as a particular hurdle by [PM2], [RI2], and [RI1]. Advancements in manufacturing environments are being introduced, however: [RI2] cited new, adjustable jigs (assemblies for attaching parts) as a means to improve flexibility (i.e., allowing easy modification in order for systems to operate in a new fashion or with new workpieces [see Section 5]) and robustness, while [RI1] reported leveraging deep learning as a means to improve gripping performance in industrial environments. [PM2] has utilized low-cost additive manufacturing (AM) technologies as a means to quickly and cheaply tailor grippers to parts, made possible by the relatively light payloads within their particular industry. [PM3] cited many manufacturers with new gripping systems, and added that these systems are becoming progressively cheaper.

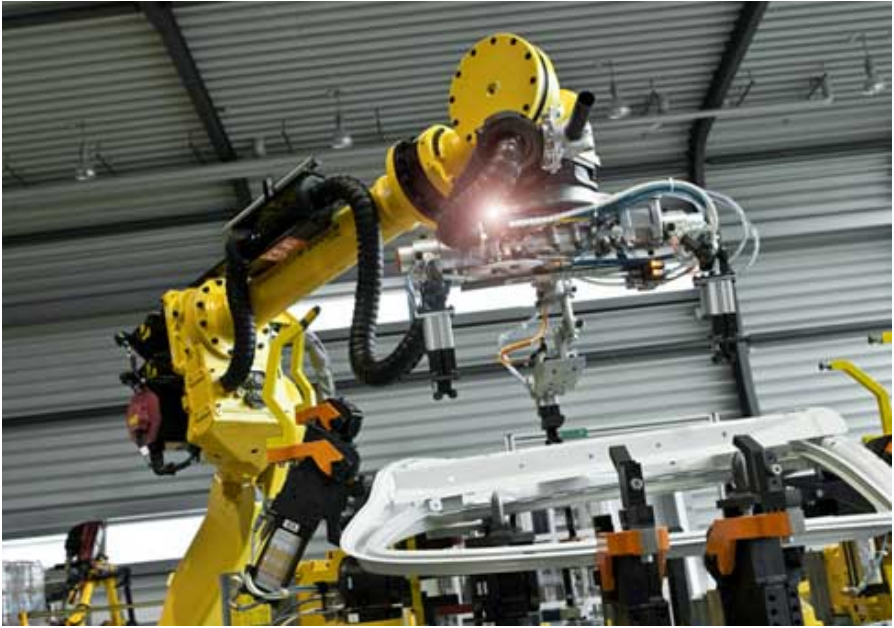

**Figure 3.5:** Industrial robot gripper (Fanuc, 2020)

Severe challenges remain, however; among them, [PM1] cited the need for better gripping hardware and more robust sensing systems, while [RM2] identified the need for safe grippers in collaborative environments. In particular, [PM1] highlighted the need for gripping hardware that will work for various workpieces without requiring modification of the workcell. Most importantly, many companies stressed the need for robustness and reliability of systems involved with gripping and manipulating workpieces. Robustness appears to have posed a barrier to innovation, with [PM1] stating that robotics could be more flexible if gripping were more flexible.

### 3.1.5 Sensing and Perception

Sensing and perception systems are designed to extract usable information from the environment. "Sensingâ"Ì generally refers to hardware sensors, while "perceptionâ"Ì refers to the software that extracts usable information from those sensors. While some companies reported that sensing systems have improved n recent years, [RI2] and [PM1] cited a need for improved systems, both to better utilize more advanced approaches (e.g., deep learning), and to improve the robot's perception of the environment.

Some advancements have posed their own challenges. [PM1] noted that deep learning in particular has been an enabling technology for automated quality control of produced parts, but has failed to provide the robustness necessary for object and pose recognition to be viable in a factory setting ([PM1] added that the performance of these technologies degrades with simple parts in such a setting, and further noted that some algorithms do not outperform template matching on the factory floor). [RI2] said that similar perception algorithms have failed to provide information at a sufficiently high rate to be useful to a high-speed robot.

In contrast, [PM3] said that sensing and software advancements (i.e. the means to intelligently process increasing amounts of sensory data) have allowed for new manufacturing capabilities in recent years, supporting the integration of autonomous guided vehicles on factory floors through advanced perception algorithms such as SLAM (simultaneous localization and mapping (Dissanayake *et al.*, 2001)).

There is a common perception among those interviewed that recent perception algorithm advancements may be helpful for the predictive maintenance of workcells through condition monitoring, the identification of key performance variables and characteristics (see section 3.2.7 for additional detail). [RI1] said AI and improved computer vision techniques have helped to improve automated quality assessment systems; [RM4] further highlighted the successful use of tactile sensing in their solutions, an avenue the company often explores before considering vision-based approaches.

### 3.1.6   Interfaces for Programming and Communication

The majority of companies interviewed indicated an increasing focus on design and development of robot interfaces with regard to improved programming methods and inter-process communication. This included interfaces at multiple levels of abstraction, at both high-level programming interfaces that control robots and enable them to perform a variety of tasks, and low-level communication interfaces that allow robots to communicate and integrate with other hardware and software. Multiple companies, including [RM1], [PM3], and [RI1], emphasized that because robot hardware is now largely identical between companies, interfaces represent a major differentiator within the market and are a constant focus of robotics companies. An example of a current robot programming interface, called the Teach Pendant, is shown in Figure 3.6. Here, we detail current company priorities in terms of easier-to-program robot and communication interfaces.

In some capacity, all interview subjects discussed easier robot programming enabled through easy-to-use robot interfaces. Multiple companies identified an increased focus on easy-to-program robots, which accompanied a resurgence of interest in collaborative robot systems that began in about 2010. [RM2], [RM3], and [PM1] said one major draw of collaborative robots at that time was the idea that they were easy to program, since many could be programmed using graphical languages (rather than languages such as Java or C++). However, opinions on whether collaborative robots actually are easier to program varied across the interviews. [PM1] stated that while they might be easier to program for simple tasks, they are more difficult to program for off-nominal or requirement-dependent systems. As a result, [PM1] has developed their own easy-to-use interface built upon existing robot programming interfaces, such that programmers only interact with a simpler, abstracted interface.

Some of the major draws of easy-to-program robots according to our interview subjects included increased flexibility, faster prototyping ([PM2]), and a reduction of the time it takes to integrate robots into manufacturing lines ([PM1], [RM3], [RI1]). [PM1] explained that robots themselves now cost very little relative to the costs of integrating and programming them. They said integration (discussed further in Section 5.1) and the process of setting up a manufacturing line are both very expensive, and reprogramming robots and adjusting processes after completing this initial setup requires companies to continue to bring in external integrators. [RM3] explained that high integration and programming costs mean that SMEs, which have less financial flexibility than more established companies, are often only able to employ robots for simple tasks. [RI1] said they consider both reprogramming and reintegration time when developing new robotic solutions as a result of such factors.

Driven by the need for easier-to-use robotic systems, a number of small startup companies, including one we visited, are focused on the ease of programming of robots. [RM4] demonstrated interest in easier development and sharing solutions between the research community and industry; they have developed a graphical programming language for easier programming, but also enabled users to program the robot with more traditional languages such as Java and C++. They also developed an online platform that allows solutions to be shared with other manufacturers.

While small startups such as this are making progress towards easier-to-use robotic systems, [PM3] mentioned that young startups may not yet be able to hasten integration by reducing programming time due to robustness requirements for these manufacturing lines that new, easy-to-program robots can not meet in all cases. Furthermore, the systems built by startups often prove difficult to integrate into existing, potentially older and proprietary IT environments in large manufacturing companies. We discuss the robustness considerations further in Section 5.5.3.

Multiple interviews also addressed communication interfaces, including robot programming languages and communication protocols for integration of robots with other hardware and software. Multiple companies ([PM3] and [RM3]) discussed the current variety of programming languages: [RM3] explained that each robot possesses its own programming language and communication protocol, with 17-20 different languages currently in use, most of which are proprietary. [PM3] stated that the Robot Operating System (ROS) (ROS, 2020) might emerge as a standard programming framework for industry due to its de facto standard use in academic research and thus widespread knowledge among the next generation of robot programmers; many companies also mentioned OPC UA (OPC Foundation, 2020) as a potential standard protocol. However, most suggested an overall need for more standardized communication interfaces, which we discuss further in Section 5.2.2.

### 3.1.7 Autonomous Guided Vehicles (AGVs)

Autonomous guided vehicles (AGVs), also referred to as "mobile robots," are categorized as service robots by the IFR and ISO (IFR, 2019; ISO, 2012). The ISO defines a mobile robot as a "robot that is able to travel under its own control" (ISO, 2012). AGVs typically have wheels and sensors that allow them to navigate around factory floors. They were discussed in seven of the nine interviews, and represent a current major area of focus. Multiple companies, including [RM1], [PM2], [PM1], and [PM3], have already implemented AGVs on their factory floors, either for demo purposes ([RM1]) or for actual production (all others); other companies mentioned that they plan to incorporate AGVs in the immediate future ([RM3], [RI2]).

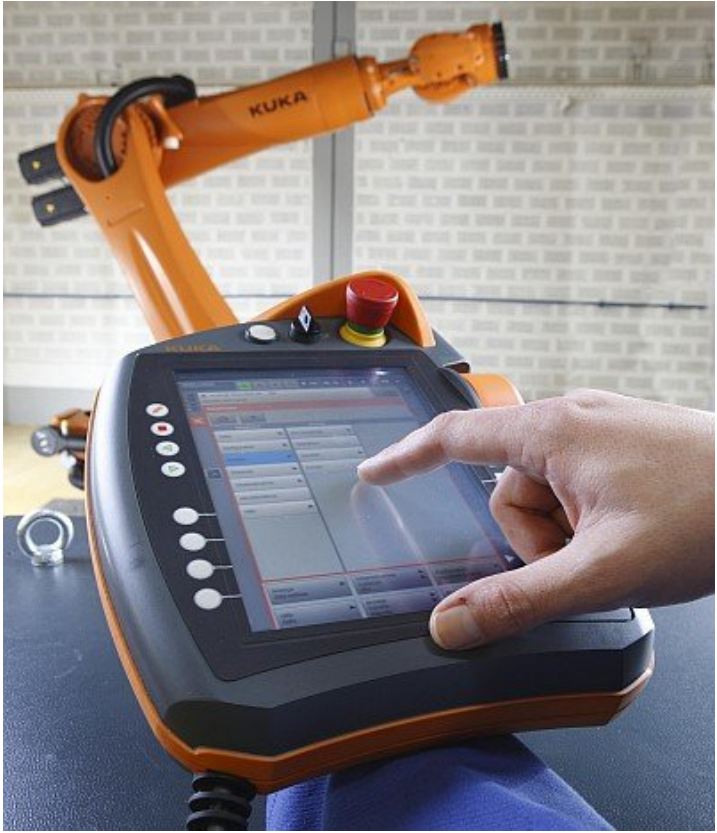

**Figure 3.6:** Teach Pendant interface for robot programming (Restrepo, 2018)

Currently, AGVs in factories are largely used for labor-intensive domains, such as logistics (Figure 3.7) or moving material between workcells on factory floors. While [PM3] emphasized how much more capable AGV technologies are today than they were 10 years ago (largely due to new sensing and algorithm capabilities such as simultaneous localization and mapping [SLAM]), [PM1] stated that the factory environment remains difficult for these systems to navigate. Beyond this, [RM3] and [PM1] also acknowledged that there are not yet standards for AGVs or fleet management of mobile platforms; [PM1] ex-plained that such standards are emerging and driven by the automotive industry, where a high demand exists for such systems.

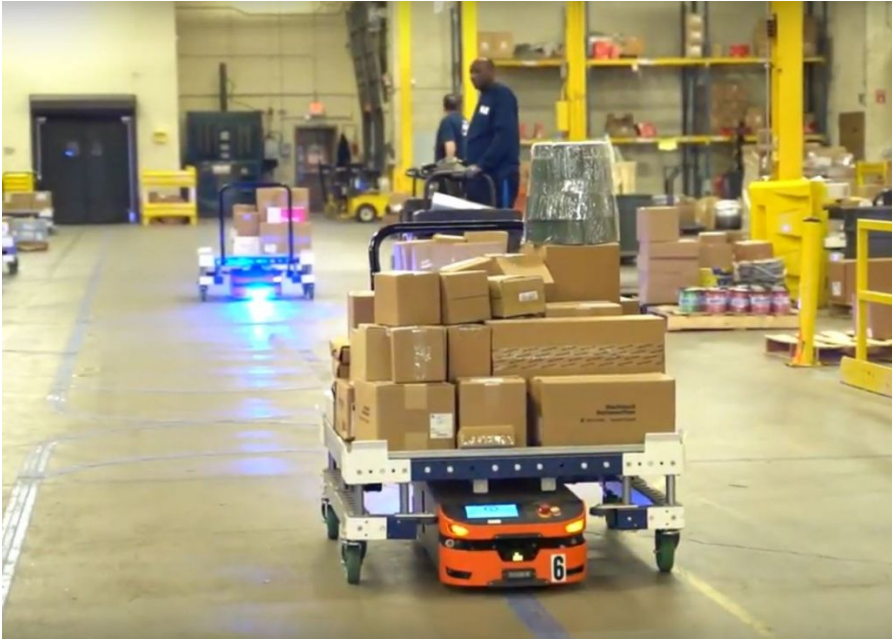

**Figure 3.7:** Autonomous Guided Vehicle (AGV) performing a logistics task in a factory (FlexQube, 2019)

## 3.2 The Internet of Things (IoT) and Computer Systems

As detailed by [RM3], [RM2], [RI2] and several others, IoT integration has been slow, hampered by data ownership and security concerns alongside an unclear value proposition. According to [RM3], "people aren't ready to open their factory up to the internet or cloud." [RM1] cited concerns about political problems in transferring information across national borders, as well as issues determining whether computation should occur on the cloud or on the edge. [RI1] stated that usage is more for information exchange and less for remote control and operation, but that the primary limitation is in data infrastructure. [RM2] and [RI2] noted that old hardware and machinery limit the ease with which IoT may be integrated into a factory - which, in the absence of a clear benefit from doing so, makes it difficult to motivate investment.

How to use the data posed a common concern. The popular value propositions stated by [PM2], [RI2], and others were that IoT may be useful for quality control, predictive maintenance, and production statistics. [RI1] also said that interconnectivity may have prospective benefits in the future. However, it appears that IoT technologies have yet to fully permeate the industrial landscape and to unfold its full potential for production management and optimization; these technologies are currently likely in a "pre-development and pilot stage" (as identified by [PM3]).

### 3.2.1   Simulation Systems

[PM2], [RM1], [RI2], and others recognized the importance of simulation in their design and development processes and emphasized an increased focus on simulation with the push for Industry 4.0. Simulation solutions discussed in interviews included both simulations of robots or other tools at a manufacturing cell level and high-level simulations of value streams in a manufacturing plant. Across all companies, current uses of simulation solutions included determining the feasibility of robotic and other manufacturing technologies, informing which robotic or other tools to purchase and implement for manufacturing lines, calculating the economic value added (EVA) for different manufacturing setups, determining the cost of robotic solutions over the robots' entire lifetime, informing human-centered design processes, quickly testing and generating robot programs at the cell level, and generating synthetic data to help with data scarcity for vision-based part inspection.

Robot manufacturers, research institutions, and product manufacturers all reported an emphasis on the uses of simulation solutions in the design and optimization of manufacturing processes. [RM3] produces their own simulation product and stated that simulation is a significant focus because it can simplify the integration of robots into manufacturing lines. [RM2] stated, "simulation is our way of life," while [RI1] and [RI2] both identified simulation as both a current and future focus. [RI1] in particular reported working on a modular robot architecture that would allow users to simulate solutions before implementing them on their manufacturing lines. [PM3]  have created a simulation of the entire value stream of their factories in order to better understand where automation could be beneficial.

[PM2] said automation and simulation represent "two sides of the same coin," and that they view it as too risky to automate in many cases without first simulating their automation solutions. They discussed simulation at length, and explained that there has been a greater democratization of simulation solutions in recent years, with line workers now monitoring and utilizing simulations in some cases. They have implemented simulations at the plant level for EVA estimation, and at the cell level to understand how robots will move and whether there would be any issues with robot operation in the given cell (such as arm collisions).

[PM1] said it has become computationally easier to make better plant and workcell simulations in the past 10 years, and that they now possess a 3D model of an entire manufacturing line. They expressed hope that simulations will eventually reduce integration times; however, because simulations do not perfectly represent reality, they do not currently achieve this purpose. (Multiple companies mentioned the potential for improving simulations by integrating real manufacturing-line data, as discussed in Section 6.3.)

Two specific types of simulation solutions discussed or demoed during the interviews included digital twins and augmented and virtual-reality systems, each of which we detail in the following sections.

### 3.2.2 Digital Twins

A "digital twin" is defined as "a real mapping of all components in the product life cycle using physical data, virtual data and interaction data between them" (Tao *et al.*, 2019). Digital twins were discussed during over half of the interviews, mentioned in conjunction with broader conversations around efforts toward Industry 4.0.

[RM3] explained that digital twins have already existed for a few years in the form of real 3D environments, and have been used to validate line performance. [RI2] reported creating digital twins to simulate tool behavior, including their material properties (using finite element methods, for example), with the eventual goal of integration of digital twins into robot interfaces and IoT integration in robot cells. Finally, [PM2] highlighted digital twins as an "enabling" technology that can support flexible automation by reducing integration times from multiple weeks to a few days, allowing companies to get "smart data from big data." They reported developing digital twins at both the

individual product and production process levels: specifically, they mentioned working toward supporting additive manufacturing processes through digital twins of related production services and tools.

### 3.2.3  Augmented Reality and Virtual Reality Systems

Virtual reality (VR) is defined as an immersive or semi-immersive interaction interface incorporating a simulation of the world that includes 3D geometry space (Ong and Nee, 2013). It often involves special hardware, including head-mounted displays and gloves (Ong and Nee, 2013), as shown in Figure 3.8. Augmented reality (AR) is a form of human-machine interaction that overlays computer-generated information onto a real-world environment (Ong and Nee, 2013); in contrast to VR, it enhances the existing environment, rather than replacing it altogether (Ong and Nee, 2013). AR also involves special equipment, which can include head-mounted displays.

Both VR and AR were mentioned during interviews, but a greater emphasis was placed on VR in terms of current and future solutions. [RM3] said they use VR as a research and development (R&D) technology for pre-integration of robots and to check the feasibility of robotic solutions. [PM2] showcased two demonstrations of current and ongoing VR projects during our visit. The first project integrated a computer-aided design (CAD) model of a product with a thermal simulation, and was used to support manufacturing design. The second project involved a virtual simulation of workcell tools and was used for assembly process design. The primary goal of this second solution was to reduce integration costs by testing solutions in a virtual environment, affording the ability to check for collisions between tools within a variety of scenarios. While the VR technologies we observed were preliminary, a number of companies discussed the potential of expanding VR capabilities in the future.

### 3.2.4  Cloud Systems

Since a major aim of Industry 4.0 is to increase IoT and cloud connectivity (Monostori *et al.*, 2016), most companies discussed cloud systems in relation to Industry 4.0. We define "cloud systemsâĂİ as those that provide computing resources over the internet for data handling, processing, or communication. While cloud computing is a goal of Industry 4.0 and most companies described

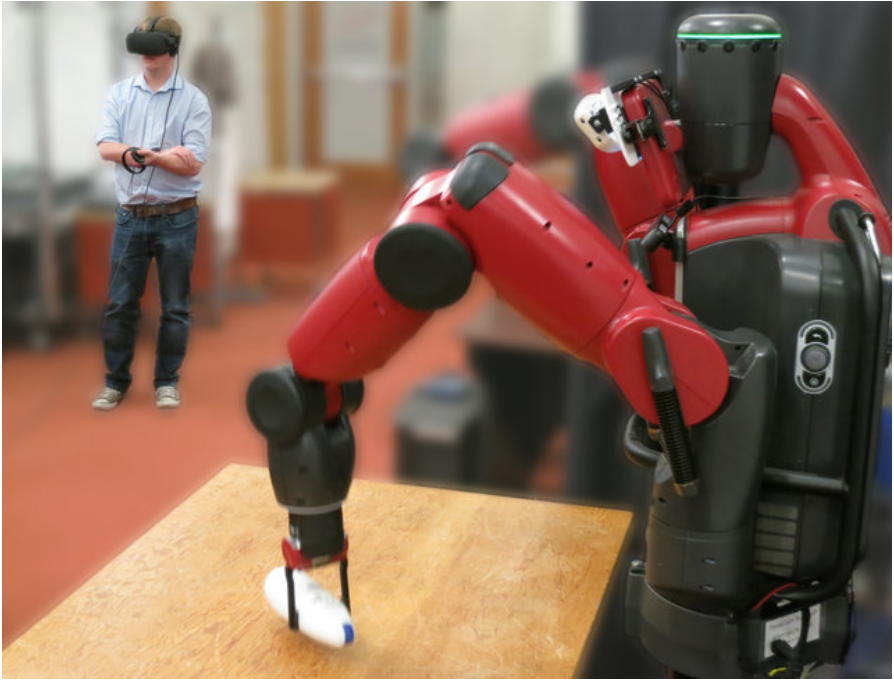

**Figure 3.8:** Prototype virtual reality system for robot teleoperation demonstrated in research (DelPreto *et al.*, 2020)

work toward this end, they also pointed out challenges associated with the use of cloud systems.

[RM3] and [PM1] acknowledged privacy concerns around connecting factories to the cloud, and the potential to have intellectual property (IP) compromised by doing so. Further, [RM1], [PM2], [PM1], and [PM3] discussed issues related to the lack of standardization of cloud systems and associated protocols for communication between manufacturing-line hardware and the cloud, which we address further in Section 5.2.2. They also remarked on how large robot, product, and automotive manufacturers often develop their own cloud platforms to meet their own needs, while smaller companies frequently use the systems developed by these larger companies.

[RM1] explained that robot manufacturers must develop their hardware to be agnostic to whichever cloud platform their customers choose; however, multiple manufacturers using cloud systems reported that integrating hardware

from many different suppliers is challenging due to the lack of standards, with additional hurdles arising because some manufacturers are using 15-20 year-old technology that is either difficult to connect to the cloud or cannot be connected at all. As a result, companies that integrate cloud systems currently develop tailored workaround solutions specific to the old hardware that they are using.

Finally, [RI1] had developed a mock factory to test cloud navigation, but also raised concerns related to the reliability of communications through cloud systems. While progress has been made toward development of these systems, issues related to standardization, security, and reliability will need to be addressed in order for this technology to realize its full potential.

### 3.2.5 Production Control

One major vision of Industry 4.0 enabled through cloud computing is more interconnected factories, allowing for centralized control and optimization of production processes. Four companies - [RM1], [RM4], [RI1], and [PM2] - mentioned production optimization and control as a current or future goal. [RM1] reported working on a production optimization platform that leveraged data collected from their robots. [RM4] said they recognize the benefit of monitoring and controlling entire lines in a centralized way, and noted the value of interconnecting robots by exchanging programs between them. [PM2] stated that they are considering production optimization using the digital twins they have under development. Finally, [RI1] said they are working toward improving the interconnectedness of systems within the cloud to enable "automation of automation," tying together individual manufacturing processes. Although this is the eventual goal, they explained that IoT and cloud systems are currently being used for information exchange only and not yet for production control.

### 3.2.6 Data Infrastructures and Data Handling

Most of the companies we interviewed mentioned data in relation to Industry 4.0. Manufacturers reported collecting data by adding sensing to their manufacturing tools, workcells, and products, with the intent to use the resulting data to improve production processes. Many companies stated that they now collect large amounts of data, but most do not yet know what to do with it.

One reported challenge is the current lack of adequate infrastructures for data handling: without adequate infrastructures, generating value in the form of usable information from raw data is difficult. ([PM2] referred to this problem as getting "smart data from big data.")

[PM1] stated that which data should exist in the cloud and which should remain on the ground remains an open question, and software solutions for connecting the cloud to the ground represent another open area of exploration. Companies also acknowledged difficulty related to a lack of standardization for data handling, which we discuss further in Section 5.2.2. While there are limits to how data can be used based on the limitations of existing data infrastructures, [PM2] said they are currently using data for predictive maintenance (as discussed in the following section [3.2.7]) and quality control. They also hope to extend their use of data for digital twin creation, which can be supported by enhanced data infrastructures.

### 3.2.7   Predictive Maintenance

Five companies highlighted predictive maintenance as a prime use for their collected data. Predictive maintenance incorporates direct data about a tool's condition in order to schedule maintenance procedures, rather than mean-time-to-failure or similar metrics (Mobley, 2002). Multiple manufacturers said they have already implemented predictive maintenance in their factories, reducing repair and replace costs for tooling in manufacturing lines. While predictive maintenance represents one current use for data, most companies said they hope to eventually expand their data usage to additional applications

# 4

---

# Company Processes

---

We asked companies which new technologies they had introduced in the last 8-10 years and the decision making processes for their introduction of these technologies. Responses indicated a spectrum of approaches, some of which seemed to allow companies to adapt more quickly than others. In this section, we report some of the main points discussed, including the variety of company mindsets around the adoption of new technologies, whether and how companies leverage worker expertise to drive innovation, and the metrics the companies considered when justifying automation.

## 4.1 Company Mindsets

Companies that we interviewed or that interviewees discussed during the interviews fell into two primary categories in terms of their approaches to introducing new automation technologies onto their manufacturing lines: those that fit tasks to their automation capabilities and those that adapted automation capabilities to meet their task requirements. We call the former a "top-down" mindset and the latter a "bottom-up" mindset. In general, companies employing a "bottom-up" mindset adopted more new technologies in recent years and were using them in more novel ways than those employing a "top-down" approach. This might have been due to the fact that the "bottom-up" approach does not tie companies to specific automation solutions, but rather leads to the implementation of only automation technologies that are useful for specific manufacturing tasks.

One example of a company with a "bottom-up" approach was [PM2] , which had implemented a "Robot Experience Center (REC)" on their factory floor. At the REC, robotics engineers work in close proximity to the manufacturing lines as they generate new ideas for how to apply robotics to different tasks along the line, prototype solutions, and change lines after they are already up and running. The engineers in the REC incorporate ideas about new automation solutions from all over the company, including line workers. They primarily use compliant, "safe" robotic arms for prototyping and on their assembly lines, allowing them to try new solutions more quickly and safely than with standard industrial arms. The company stated that the economic value added (EVA) must be proved out in the REC before a solution is actually implemented on a line.

[RM4] emphasized the challenge of integrating new technologies, such as compliant arms, in companies with outdated ("top-down") mindsets with regard to robotics technologies. They gave an example of a German automobile manufacturing company employing a torque-controlled robot arm as though it were a position-controlled system while only using the compliance capabilities for sensing people. This approach, they said, misses the point of these compliant robotic arms, because much more can be done with them. [RM4] reported difficulties working with older, larger companies, which typically have mindsets about the way in which robots should be used that limit their ability to identify the "right" tasks to automate. They said younger, smaller companies "immediately choose the right tasks to automate" because they have less financial flexibility and are focused only on solutions that directly meet their needs. [RM4] aims not to sell its products for use in classic automation settings, since this often limits how its robotic arms are used.

Four separate companies, including [RM1], [RM4], [RM2], and [PM2] independently discussed the idea of "programming the task and not the robot" as an important next step in terms of technological advancement. [RM4] referred to this strategy as "empowering people to empower robots." The majority of interview subjects also addressed the idea of making easier-to-program robotic systems to allow workers to directly program or reprogram a robot to perform specific tasks. Overall, this would enable companies to more easily employ "bottom-up" strategies and could yield greater flexibility in terms of what robotic systems can achieve in different manufacturing contexts as well as how these systems are used on manufacturing lines.

## 4.2  Leveraging Worker Expertise to Drive Innovation

A common theme across many of the interviews was innovation enabled by leveraging factory worker expertise. [RI1] , which has interacted with a number of industry partners, explained that many of the "most impressive robot solutions" they had seen were those that involved factory workers in commissioning a product through a co-creation process. They stated that SMEs were more effectively implementing co-creation processes than larger companies and that larger companies suffered negative consequences (such as workers' fear or dislike of the robots) as a result.

Beyond this, "lights out" factories, or those that are fully automated with no human workers during nominal operations, were discussed in many of our interviews. While almost all subjects sais they did not see lights out factories as a possibility in the near future for a variety of reasons (because it is not cost effective, certain tasks remain difficult to automate, the need for a human presence in the event of an emergency or other problem, etc.), [PM1] in particular emphasized the value of human labor as a source of innovation: "Improvement in the factories stops with lights out factories, because line workers are continuously telling us how processes can be made better." Overall, the value of worker domain knowledge in the innovation and implementation of new robotic solutions was a frequent topic of discussion.

[RM4] said most of their customers are experts in their fields and that the company's robotic solutions are tailored to those customers' specific applications. In order to enable customers to share solutions that require specific domain knowledge, [RM4] developed an online platform they described as an "app store for robots," which allows users to develop and share task-specific solutions without the company's direct involvement. They also opened this platform to the research community, with the hope that it would enable the efficient transfer of knowledge from researchers to the industry.

## 4.3 Metrics

We asked companies about the metrics they use when deciding whether to automate, what to automate, which technologies to use when automating, and the benefits of automation. By far, the most common responses were related to costs, with primary drivers including labor costs, transportation and tax-related costs linked to non-collocation of production facilities, and integration costs (both in terms of ramp-up time for manufacturing lines and direct costs for hiring external integrators). With respect to labor costs, [PM3] said labor displacement does not necessarily make sense with new collaborative systems, since true collaboration does not always imply a reduction in labor costs; a number of companies discussed total cost of ownership (TCO) or total return on investment (ROI) calculations over the lifetime of technologies as a more accurate tool for innovation management, with calculations facilitated and enhanced via simulation.

Companies also mentioned productivity, quality, ergonomics, labor shortages, precision, and reducing manufacturing line footprints as reasons to automate, with many citing an aging workforce as a cause for increased focus on ergonomics in robotic solutions. Further, while a number of companies expressed interest in reducing factory footprints, [PM3] placed particular emphasis on being able to perform more non-value added tasks within a smaller footprint. They noted that with new collaborative systems and fewer cages around robots, reducing footprints (both physically and financially) is becoming increasingly achievable. Finally, multiple companies mentioned flexibility as an important reason to automate. [PM1] listed the flexibility, versatility, and reconfigurabiltiy of manufacturing lines as secondary metrics for automation (after cost). However, [PM3] mentioned that although flexibility can have many secondary benefits, such as reusability of robotic systems and the ability to manage increased variation in product lines, it is inherently difficult to quantify and therefore to consider in a profitability analysis, since there is not an agreed-upon definition.

## 4.4 Line Workers' Role in Manufacturing

In our interviews, we asked about the role of the line worker in manufacturing with regard to the introduction of new technologies. While we were told there are an increasing number of collaborative robot applications and fewer cages around robots, companies also emphasized that there remain tasks robots can perform behind cages that do not require human involvement, such as welding and other activities along an automotive manufacturing line (Most of these applications have been automated for years already.). Conversely, some tasks still require or are more cost-effective to perform with people, such as moving material between workcells, cleaning machines, performing maintenance, and troubleshooting, along with many non-value-added tasks during final assembly processes. [RM4] said there are not many truly collaborative applications for robots and that, for example, the number of direct interactions humans have with each other on manufacturing lines is very small. [RM4] believes the primary value of robots is their ability to perform work without having to wait for a human and are more focused on human-robot coexistence and robot teaching scenarios than explicit collaboration. In short, they view robots as "power tools" that can enhance human capabilities.

[PM3] explained that they still leverage human cognition for flexibility and primarily view robots as providing assistance to humans, particularly during the final assembly process. Multiple companies also highlighted the a shift toward more highly-skilled work, such as robot programming and maintenance. [PM2] noted that people increasingly need to deal with software and new technologies in manufacturing, so the company supports ongoing education and upskilling of workers at all levels.

# 5

# **Challenges for Introduction of New Technologies**

In addition to asking about technologies the companies had introduced and the decision making processes behind introducing these technologies, we also asked about the primary challenges companies face when implementing new robotics solutions. Seven primary categories emerged in response: integration, standardization, flexibility, skills, education and training, technological bottlenecks, and social and ethical considerations. These are detailed in the following sections.

## 5.1 Integration

A common major theme throughout the interviews was the challenge of integrating robotics into manufacturing lines. In the manufacturing ecosystem, companies often draw upon system integrators to design assembly lines. Many large robotics manufacturers are also system integrators and incorporate their own hardware into integrated solutions in addition to that of other companies. Independent system integrators also exist, and many small integrators work on specific manufacturing applications (e.g., welding, painting, etc.). Most companies we spoke with discussed the challenge of integration time and associated integration costs. [RM1] stated that robots are now inexpensive, but integration is not (in fact, integration accounts for 4-5 times the cost of the actual robot). Throughout all interviews, the interviewees explained that many manufacturers currently experience extended ramp up times as they iterate with integrators to adjust assembly lines until they function properly. Beyond this, if small changes must be made to the line, or if workcell components break, manufacturers must continue to draw upon system integrators to make modifications. This causes inflexibility along manufacturing lines (which are often carefully set up to perform very specific tasks) and and inability to reuse the robots employed on those lines. [PM3] said that it often proves more economical to dispose of all robots used on one manufacturing line once they are dismounted from that line for this reason, but that they hope for increased robot reusability in the future.

Beyond the challenge of setting up and maintaining manufacturing lines with system integrators, a number of companies also mentioned the scarcity of integrators within the robotics ecosystem. Some companies reported encountering additional integration challenges because not enough integrators exist to meet the demand. [PM2] viewed their reliance on integrators as a liability and decided to bring integration in-house. The company explained that they did not have a background in integration and that they made several mistakes when they initially began to perform integration work. The company viewed their initial slow integration pace as worth the temporary additional cost, however, given the long-term value value of knowing how to perform integration within the company. [PM2] now almost exclusively uses safe compliant robotic arms for integrated solutions, because they have learned it is easier and faster for engineers to prototype and test solutions in close proximity to robots without worrying about human/robot collisions.

[RM1] explained that one of their primary focuses as a company is to reduce integration costs by developing systems that are easy to program and re-program. They also emphasized the costs associated with "re-integration" (costs that arise when small changes must be made to existing lines) and the need to allow people within factories to make these minor adjustments on their own. While many of the companies mentioned ease-of-programming as a possible enabler of reduced integration times, [PM3] said that while it is one possible prerequisite, ease-of-programming alone may not necessarily reduce integration time, especially for robustness-reliant industries like automotive manufacturing. They stated that some of the new, "easy-to-program" systems are not making integration faster yet, because the robot arms they are using

lack the industrial-level robustness they require.

## 5.2 Standardization

We asked companies about the state of standardization of hardware, software, communication, and safety requirements and other potential areas of standardization more generally. One key theme that emerged in their responses was that development of many types of standards including for hardware, software, communication, and safety, is still in progress, which limits companies' ability to adopt new technologies. Lack of standards increases integration time and cost and reduces the reusability of robotic systems for manufacturing multiple types of products or for use on new assembly lines. According to [PM1], "robots are relatively cheap today, so many companies will just get rid of them rather than reuse them when a line is done."

Eventual development of common standards across the industry could improve flexibility and system modularity, which in turn could increase uptake of robotic technologies. [PM1] said, "We have to standardize, because lot sizes are decreasing, and production amounts are decreasing. We need standardization to enable versatile production." However, they added, "Standardization is not as easy as one thinks; there are many experts to listen to. It must go hand-in-hand with which technologies are capable of what things and an understanding of the processes they're a part of. It mustn't cost too much money."

Across all interviews, safety standards, communication protocols and standardized interfaces, robot programming languages, hardware, and robot specifications were cited as bottlenecks for the integration of new technologies into manufacturing. We detail each further in the following sections.

### 5.2.1 Safety Standards

One theme that emerged in a number of interviews was the challenge of defining safety standards, especially as they pertain to new, collaborative robot systems. Many companies acknowledged a confusion around how to define safety standards for collaborative and cageless robots. A number of companies mentioned the ISO 15066 standard (ISO, 2016), which provides some safety guidance for collaborative systems; however, they noted that current standards primarily revolve around velocity and force limits, which may not be sufficiently comprehensive to ensure safety of nearby human workers. In the factories we toured, we observed both compliance-based approaches to safety and robotic systems with capacitive skins for contactless human detection, two disparate approaches to safety that each merit distinct considerations. [RI2] said they believe companies must be involved in the definition of safety standards, since they are the ones using the new, collaborative systems.

Multiple companies independently discussed the idea that robotic arms may be safe, but the tasks they work on or tools they use may be unsafe [RM1] said there was a lot of confusion in the industry when robot arms began to be marketed as "safe" and that companies purchasing these arms had mismatched expectations since, in order to ensure safety, the applications the robots were working on had to be safe in addition to the arm itself. They explained that system integrators were ultimately responsible for safety certification and acquiring CE Markings (which indicate that products meet European Union safety, health, and environmental protection requirements - a necessity for any product marketed in the EU (*CE Marking* 2020)) for particular system set-ups, but many companies did not realize this when robotic arms were first marketed as "safe". Multiple companies we interviewed stated that as a result, collaborative robotics is essentially synonymous with "slow-motion robotics" at this point, and that the industry is not yet able to use these technologies to their fullest potential. We observed one key example of this during a tour of [PM2] , which almost exclusively used collaborative arms for manufacturing, but with all of the arms behind Plexiglass on manufacturing lines, because the parts the robots were manipulating had sharp edges, and it was unsafe for humans to occupy the same workspace.

[RM2] and [RM4] are taking steps to address some of these issues. [RM2] reported working on outfitting tools and work pieces/parts with capacitive skins to ensure the entire manufacturing set up was safe for humans. [RM4] said they were in the process of having their robot arms safety certified (with CE Markings) out of the box, so that an integrator would not have to be responsible for acquiring certifications (although they did not specify how this could be done).

### 5.2.2   Standardized Infrastructures and Communication Protocols

Most of the companies we interviewed discussed the standardization of cloud platforms and infrastructures in addition to communication protocols for hardware. We learned there are a few primary cloud platforms used within the industry (mainly developed by large manufacturers), but that infrastructures are generally not standardized. [RM1] explained that they have to be able to adapt to whatever customers decide they want to use and accommodate any type of platform. They stated that some large companies (such as one of their automotive manufacturing partners) make their own platforms, but many smaller companies cannot afford to develop their own solutions. This robot manufacturer had developed a platform for those small companies as well. [PM3] mentioned that all of the platform developers are independently working to enable compatibility with 15-20 year-old technology as well, since many companies are still using old hardware. In general, most companies felt that integration would be easier with a standard platform, but [RM2] explained that robot suppliers do not want to standardize, and the relative strengths of different actors within the robotics ecosystem would likely prevent industry-wide standardization from happening in the near future. [PM2] said they believe the lack of standardization poses a big problem and that large companies must develop standards that actually work with the hardware.

In terms of communication protocols, [RM3] reported that there are 17-20 different protocols currently used in the industry, and that each programmable logic controller (PLC) manufacturer has their own. Many companies mentioned that while machine-to-machine communication protocols are not standardized everywhere, the OPC Unified Architecture (OPC UA) may be adopted as the industry standard. Over half of the companies we interviewed mentioned OPC UA, and most robotics manufacturers have plans to support OPC UA integration.

### 5.2.3   Standardized Interfaces and Robot Programming Languages

Multiple companies discussed the lack of standardized interfaces and robot programming languages as a barrier to integration. [RM1] mentioned that interfaces represent a constant challenge, because hardware is now largely the same between robot manufacturers, and interfaces are what differentiate companies from one another. They explained that end users want standardized interfaces, but that it is difficult for robot manufacturers to justify working toward this end, as interfaces represent a potential unique selling point (USP). [PM1] is developing their own interfaces that abstract away low-level robot interfaces and programming languages such that line workers never have to deal with them directly. They said the industry would benefit from standards for interfaces and programming languages, but that a shift toward those would pose a challenge, because they have already put a significant amount of money and resources into developing their own systems which work well and are reliable. They added there would need to be an external push to adopt a new standard, because they would not choose to change deviate from their current systems otherwise.

Many companies also mentioned the Robot Operating System (ROS) (ROS, 2020) as a potential industry standard. While many were excited about the potential of ROS, especially in terms of more fluently connecting research to industry, [PM3] mentioned that ROS does not yet support the robustness necessary for industrial manufacturing. In addition, there is need for maintenance and shop floor staff to acquire the necessary know-how with regards to working with ROS-based systems, a need to be addressed on a larger scale, namely with the production system as a whole. [PM3] expressed hope that ROS 2 will be usable for more industrial applications, and that it will cultivate more system modularity in robotic manufacturing applications.

### 5.2.4 Standardized Hardware

[PM1], [PM2], and [RM1] discussed robot hardware standardization. [RM1] mentioned that in terms of hardware, flanges are standardized between systems, but not much else. [PM1] said complex mechatronic components are now being used in robotics, making integration difficult in the absence of standardization. They believe standardized hardware (including hardware interfaces) in addition to software is important for the industry. Finally, [PM2] reported already working toward standardizing hardware in their robot cells, with the aim of sharing solutions between cells within a manufacturing line, between lines within the manufacturing plant, and also across manufacturing plants within their global network. Their goal is to reduce integration and line worker training time by teaching workers to use systems in one place and then be able to work with those systems anywhere.

### 5.2.5 Standardized Robot Specifications

[RM4] said the lack of standardization in the definition of robot capabilities and the composition of data sheets is a current challenge for manufacturers hoping to integrate robots into their production processes, because it is difficult for customers to decide between systems. While some metrics (like precision) are standardized, they explained, the definitions of metrics related to newer lightweight and compliant arms (such as payload and sensitivity) remain vague and inconsistently defined within the industry. In response to this, the company added sensitivity to their data sheet with the hope that the rest of the industry would begin to converge on definitions.

## 5.3 Flexibility

One challenge that emerged as a common theme across all interviews was the inflexibility of current robotic systems in terms of the number of possible different use cases for robots and the difficulty associated with re-purposing robots for new tasks. All companies identified the flexibility of robotics technologies, including development of adaptable and dynamic robotic manufacturing solutions, as an important direction for future work. Across the companies interviewed, flexibility was considered important for increasing the applicability of robotics to high-mix, low-volume production (as discussed in five interviews); faster integration and re-integration times (five interviews); re-configurable workcells and manufacturing lines (three interviews); reducing the total required factory footprint by allowing manufacturing of multiple products along a single line (two interviews); and enabling the reusability of robotic systems (two interviews). Beyond this, several companies said greater flexibility would make robotics technologies more readily usable by small- and medium-sized enterprises (SMEs), which often work in high-mix, low-volume production and which make up 60-70% of all manufacturing jobs in OECD countries (OECD, 1997).

[PM1] said it does not currently make sense to automate production unless a manufacturer plans to produce a large number of items, because ramp-up and integration times for manufacturing lines are long. [PM3] and [PM2] discussed the difficulty of adding additional automation once a line is already up and running - something both would like to be able to do, but would require

more flexible systems. Finally, [PM3] said the value-added segments of their line, such as final assembly, require greater flexibility and are therefore less automated currently; they would eventually like to increase automation of these value-added segments of their manufacturing lines.

[PM3], while also expressing interest in increased flexibility, stated that "flexibilityâĂİ itself is difficult to quantify. They defined flexibility in automation as "having the opportunity to introduce new processes/updates to an existing line at low costs, for example, through a robot that can handle multiple parts and can be easily reprogrammed for new tasks," but also noted that there are many ways to think about the concept. They added that automation should always support a lean process, shifting importance to prior lean process optimization in order to identify the core added value of automation. Although flexibility might mean different things to different companies, those we interviewed mentioned several key enablers of increased flexibility, including ease-of-use/ease-of-programming, intuitive interfaces, system modularity, and simulation.

## 5.4  Skills, Education, and Training

[RM2] said that human capital is an inhibiting factor in the use of new technologies - that technological skills are lacking among workers. This has also been a burden for emerging robotics developers, who have invested in educational seminars with SMEs and the broader community, including schools. [PM2]  also detailed how they increasingly require highly-skilled workers who can operate new technologies, while [RM3] emphasized creativity as an increasingly important attribute among workers as well, since creativity is a skill that robotics cannot provide.

However, the introduction of new technologies appears to demand education within the manufacturing industry about how they can best be used. [RM3] stated that many clients had bought lightweight robotic products without knowing what to do with them, and that the challenge, therefore, is to educate the market. In particular, several interview subjects reported a misunderstanding of the definition of human-robot collaboration on the part of manufacturers and the possibilities that such technologies can offer.

## 5.5 Technological Bottlenecks

### 5.5.1 Vision, Perception, and Sensing

Multiple companies considered vision technologies promising for multiple applications, including enhancing and reducing quality check costs, ensuring the safety of collaborative robot systems, and enabling generalized part selection, among others. However, many said that current vision capabilities represent a bottleneck in terms of implementing these solutions. While vision is already used for quality checks in some industrial settings, multiple companies stated that the technologies applied in those settings could be improved.

[RM4] said they always attempt to solve problems using tactile sensing before employing vision technologies, because many challenges around data and robustness remain for vision-based solutions. [PM1] described an application of a vision-based system they had implemented that broke down when small changes were made to parts that were being handled on their manufacturing line. They said vision systems pose an integration challenge because if a supplier changes something, or if minor changes in lighting occur within the factory, the entire system will often break down. Beyond this, [PM1] said that while they have used deep learning for a number of vision applications, camera systems and neural networks currently require large amounts of training data, the associated deep learning-based solutions are not yet generalizable, and they do not yet offer the reliability and robustness required for application in industrial environments. [RI2] said intelligent robotic perception does not yet exist, and that vision systems with capabilities similar to humans will be required in order to implement these technologies for some of the new applications the manufacturing industry is considering.

### 5.5.2 Gripping

In our interviews, robotic gripping was identified as another technological bottleneck for the implementation of robotic solutions. Multiple companies cited a need for more flexible grippers that can handle a diverse set of parts. [PM1] explained that most robotic manipulators remain inflexible and that robotic gripping capabilities are still far from human gripping capabilities. They added that when new objects must be handled on a manufacturing line, companies often need to completely redesign workcells around the changing robotic grippers. They also said it is difficult for current grippers to manipulate flexible, deformable objects, which limits the types of manufacturing they can be used for. [PM1] also reported problems related to parts slipping out of grippers and errors in sensitive manipulation applications. Overall, the company expressed a need for more flexible and generalizable gripper technologies.

### 5.5.3 Robustness and Reliability of Technologies

Several companies we interviewed expressed a need for extensive reliability when introducing technologies onto a manufacturing line. Anomalies or variations in manufacturing processes can represent a hurdle: lighting changes, for example, can affect perception algorithms, while anomalies in the orientation or size of a part can limit gripping capabilities; both can bring a manufacturing line to a standstill.

Any form of downtime in a manufacturing environment poses a critical concern. As explained by [PM1], "the production world is so optimized that every minute that a robot is not moving is money out the window.âĂĬ This limits the uptake of emerging technologies within the manufacturing industry: while the technology itself is desirable, the longevity and robustness of that technology is unclear. [PM3] emphasized the high requirements towards automation in automotive manufacturing, explaining that an integrated robot must run for roughly 24 hours a day, 7 days a week for at least 7 years; the company's ability to utilize the capabilities of robots produced by new and emerging robot manufacturers is therefore bounded by the perceived robustness of the incoming technology.

### 5.6 Social and Ethical Considerations

While many companies emphasized the benefits new robotics technologies can provide for the human workers interacting with them, such as assistance with task completion, improved ergonomics of workspaces, creation of new and different jobs (robot programming, maintenance, etc.), and extended manufacturing capabilities through the leveraging of relative human and robot strengths, introducing new technologies can also have negative impacts on workers, which are important to consider as well. For example, the companies discussed worker fears related to being in close proximity to robots, the robots' impact on worker social interactions and team dynamics, and potential tracking of worker mistakes using sensor data, as well as discomfort with the pace set by robots on the manufacturing line and a general discomfort related to changes brought about by the robots' presence.

Companies also noted that since overall cost is an important metric, labor displacement calculations are still performed when integrating new robotics technologies. Additionally, some companies have already reduced the size of their workforce. These potential negative impacts should be assessed at the design phase while considering that design choices can impact whether robots are used to extend human capabilities or replace humans altogether. We acknowledge that we did not interview line workers in our study, which represents a limitation of this work. Consideration of the line worker perspective, drawing upon findings from studies such as those by Elprama *et al.* (2016), Elprama *et al.* (2017), Sauppé and Mutlu (2015), Welfare *et al.* (2019), and Wurhofer *et al.* (2018), will be critical for the design of new robotics technologies - both to ensure successful integration of technologies and for ethical purposes.

# 6

---

# Company Directions

---

Based on our findings about current and emerging technologies (Section 3), company decisions and processes around the introduction of new robotics technologies (Section 4), challenges related to the introduction of new technologies (Section 5), and current company focuses as outlined in the interviews, we now detail themes related to future directions for the companies included in this work.

## 6.1  Ease-of-Use/Ease-of-Programming and Intuitive Interfaces

The majority of companies we interviewed stressed the importance of increasing robots' ease-of-use and ease-of programming. Many reported a goal of eventually enabling line workers to directly reprogram robots without the need for extensive software or programming knowledge; this would allow workers to leverage their deep domain knowledge to rapidly make small modifications to manufacturing processes, either for improvement of these processes or to meet updated production needs. A majority of the companies stated that improved ease-of-use would also enable faster solutions prototyping for manufacturing lines, address integrator scarcity in the ecosystem, reduce teaching time for robots, make the simulation-to-programming pipeline more efficient, and increase the flexibility and reusability of robotic systems (such that it would no longer make more economic sense for companies to scrap all robots after deconstructing a manufacturing line).

[PM1] also discussed how ease-of-use could significantly impact the ability of small- and medium-sized enterprises (SMEs) - which, they said, will likely drive the market for collaborative robots - to use robotic technologies. [PM1] explained that SMEs lack the internal competence of large companies in terms of programming robots and might not have access to as many engineers. Further, they usually produce a smaller volume but greater variety of products; it will therefore be important for them to have robotic systems that can be easily set up to perform different tasks.

A number of companies mentioned the idea of "programming the task instead of the robot" as a desirable direction with regard to the ease-of-use of robotic systems, but different companies had different conceptions of what this might look like. Some discussed the promise of graphical programming languages for robots to this end. [PM3] suggested that, in the future, line workers might not program the robots, but they might interact with them in new ways, such as showing them what to do or assisting them in other ways. [PM1] envisioned semantic-level interfaces for programming and new ways of teaching robots, including learning from demonstration and kinesthetic teaching, as discussed further in Section 7.5. [RM4] indicated a company-wide focus on ease-of-use for their robotic arms, stating "the best technology cannot help you if you are not able to use it." They have developed an online platform for sharing solutions for their robotic arms within the community (an "app store"-like platform) and allow users to program solutions using either the company's own graphical language or a lower-level programming language such as Java or C++. They also host classes to teach people at all levels how to use their robots; their overall goal is to make it easy and convenient for users to have the resources necessary to develop new solutions whether or not they have programming experience.

## 6.2   System Modularity

Many companies discussed a desire for increased modularity for both software and hardware architectures in robotic systems. They explained that modularity and plug-and-produce components would increase the interchangeability of components on manufacturing lines (leading to reduced down times), faster in-tegration times, and additional interconnectedness between factory processes. A number of companies were particularly focused on interconnections between cells on manufacturing lines. [RI1] said one of their areas of interest is "automation of automation" and solving the "island problem" of workcells operating effectively on their own but not being efficiently or easily connected. [RM4] reported a goal of interconnectedness through their online platform for sharing solutions related to their robotic arm. The company's focus is on interconnecting robots and exchanging solutions between them, enabling additional scalability.

## 6.3   Simulation

Simulation was also a key area of interest for a number of companies. [PM2] stated that "simulation is the new PowerPoint," suggesting the ability to work with simulations will be ubiquitous in the future. Most companies expressed the belief that simulations will reduce integration time, support flexible automation, and enable production optimization that reduces overall down time on manufacturing lines. In particular, multiple companies mentioned development of more accurate simulations and better "digital twins" as a next step (as discussed in Section 4).

While a number of companies already use simulations of processes to assess what to automate and how to implement robotic manufacturing solutions, many are focused on improving simulations through data integration. For example, [PM1] sees digital twins as a potential tool for error-checking, and [RI2] cited better failure prediction as a potential benefit of improved digital twins. [PM2] said they hope to plan and generate robotic arm motions for the robots in their workcells via simulation software in the future instead of explicitly programming the arms. They also said they hope to integrate low-level manufacturing process simulations with plant-level simulations in order to improve the optimization of production processes. They envision integrating actual data from their lines, such as cycle times for different steps in a process, into their simulations and having these simulations dynamically impact process optimization as well. Overall, [PM2] stated that they believe simulation will be a commodity in the future.

## 6.4  Autonomous Guided Vehicles

While a number of companies have already integrated AGVs on their manufacturing floors for material movement and logistics, many said that open challenges remain in terms of their wider application and adoption. [PM1] said a high demand exists for AGVs, especially in the automotive industry and labor-intensive domains, and that they expect this demand to increase in the coming years. Across all companies, transporting materials between workcells, improving workcell interconnectedness, reducing integration times, and enabling workcell reconfigurability were all listed as benefits of AGVs.

One issue multiple companies identified as an open problem is combining mobile bases (AGVs) with robot arms to create mobile manipulators. [PM1] views the AGV navigation problem as mostly solved, but noted that the handling and manipulation problem remains difficult. Beyond this, based on the companies' statements about standardization (as discussed in Section 5),the authors note that development of standards for AGV integration with robotic manipulators and cloud systems will be critical to enabling the expansion of use cases and uptake of these systems in factories. Overall, most companies reported strong interest in AGVs and mobile platforms; further development of these systems will likely be a major next step for automation for in manufacturing.

## 6.5  Cloud Systems and Data Infrastructures

Across all the interviews, companies said they hoped to be able to use data in new ways, such as for production control and improved predictive maintenance at the plant level and load balancing between manufacturing plants at the supply-chain network level, but that currently there are insufficient data infrastructures in place to support this. The majority of companies stated that they hope for greater connectedness of robotic solutions over the cloud, centralized control of production processes over the cloud (including reconfigurability of manufacturing workcells), sharing solutions between robots, and steps taken toward "automation of automation" in general. However, they also cited a need for improved cloud-based and data handling solutions, along with greater standardization of each, in order to accomplish this vision.

## 6.6 Explainable Systems

[PM1] said that as more deep learning- and data-based solutions are implemented within manufacturing environments, it will become increasingly important for systems to be transparent. Therefore, they are considering explainable AI (XAI) as a future area of focus. XAI is defined by Gunning and Aha (2019) as "AI systems that can explain their rationale to a human user, characterize their strengths and weaknesses, and convey an understanding of how they will behave in the future." [PM1] said XAI will be particularly important for explaining decisions made by robots and AI systems when working closely with humans. The company also sees it as potentially beneficial for safety certification of perception-based systems that often leverage deep-learning techniques. While XAI was only discussed in this one interview, we view it as a potentially important focus more broadly, especially for companies that intend to expand their use of learning techniques in data-based applications.

## 6.7 Perception, Gripping, and Bin Picking

Across all interviews, companies identified important future steps for improving vision-based solutions and working toward flexible and robust gripping solutions. These were also mentioned in the context of solving the "bin-picking problem,â Ǐ which many companies identified as an important problem for improving manufacturing. As discussed in Section 5, vision-based solutions are currently limited in terms of their speed, robustness, and data efficiency. Many companies are focused on developing more generalizable vision-based solutions, which are primarily centered around deep learning.

[PM1] is working on integrating object and pose recognition into their vision pipelines, drawing on research related to that performed by Tremblay *et al.* (2018), but acknowledged that robustness remains an issue. They and others also expressed interest in improving data efficiency, and stated that the lack of transparency and explainability of these systems will also need to be addressed, especially if they are to be used for safety solutions.

Based on the gripping limitations described in Section 5, companies also discussed improving manipulation capabilities. Many are considering AI- and ML-based solutions for manipulation and for improving the capabilities of current grippers. [RI1] acknowledged solutions such as Dex-Net (Mahler *et al.*, 2017), which uses pre-trained neural networks to detect grasp points, as having the potential to enable more generalized solutions. A number of companies also discussed dexterous in-hand manipulation as a promising research direction for solving grip-related problems.

Further, over half of the companies identified generalized bin-picking as an important application they are working toward presently and will continue to focus on in the coming years. This was viewed as important for manufacturing in a general sense, but was addressed specifically in the context of the automotive assembly line in two interviews: [PM2] stated that generalized bin picking is important for eliminating excess inlay material (which they said is wasteful and expensive and is currently required to add structure to the gripping problem) in their part boxes; [RI2] said sensor-based part recognition to enable picking of parts placed in an unstructured manner is a primary concern.

# 7

# Key Research Directions

Based on our findings about the current challenges facing the companies interviewed, and the directions that they are pursuing, we now enumerate key directions for the research community to pursue that could contribute to increased uptake of robotics technologies in manufacturing environments: perception, intelligent gripping, collaborative robots, AGVs, interfaces and programming, simulation, worker-centered design, and cloud systems.

## 7.1 Perception

Perception was highlighted in the interviews as a key limitation to both the robustness of existing systems and the downstream effects on systems dependent upon those systems (such as grasping). While perception limitations were primarily reported in the context of camera systems, a broader need was identified in the demand for instrumentation to reflect factory operations in digital twins.

Vision-based perception systems have garnered much attention within academia,. While pioneering work in convolutional networks took place in the late 1990s (LeCun, Bengio, *et al.*, 1995), their use in recent deep-learning frameworks has led to unprecedented image recognition capabilities (Krizhevsky *et al.*, 2012; He *et al.*, 2015). Recognition via camera-based depth sensing has also seen improvements through deep learning: while early object-detection capabilities incorporated template-matching algorithms such as Point-Pair Features (PPF)  (Drost *et al.*, 2010), recent developments such as PointNet (Qi *et al.*, 2017) have focused on deep-learning architectures to exceed the capabilities of current state-of-the-art techniques.

However, as discussed in Section 3, deep learning-based approaches haven't delivered well on their promises in industrial environments. Our interviews revealed several bottlenecks: first, most deep-learning approaches require vast amounts of data - beyond what is typically available in a factory setting. Second, the interviewees described difficulties with adapting trained systems to new contexts (such as new workpieces). Finally, in a similar context to the data problem, the algorithms should be robust to environmental variations (such as changes in lighting).

Some of these challenges have drawn attention in the academic field. Transfer learning, or porting a pre-trained model to a new context (Tan *et al.*, 2018; Pan and Yang, 2009), appears to be of particular relevance to the data and adaptability problems encountered with deep networks. Some methods, such as the input gradient regularization method described by  Ross and Doshi-Velez (2018), aim to improve the robustness of a network to external variations, with potential relevance to accommodating environmental variations. However, it is clear that more research and development is required to fit industry needs, and accommodating new parts and minimizing data requirements for training or transferring a neural network represent valuable directions for future research. (These challenges also have a direct impact on gripping, which is discussed in the following subsection.)

In contrast to camera systems, challenges in perception for digital twins relate to large-scale representations of the active environment in a series of workcells and highlighting any anomalies. Emulation of multiple workcells may be outside the scope of the typical academic environment; however, assessing and reflecting single workcell operations may be a salient research direction. This relates closely to challenges in cloud systems, as numerous interview subjects cited difficulty making sense of the data they obtained from their factory operations.

## 7.2   Intelligent Gripping

Multiple interviewees noted how intelligent gripping has remained a significant challenge despite recent advances in gripping hardware and algorithms. Many challenges, such as maneuvering objects in constrained environments, or accommodating variations in the size and pose of a part, require a high degree of robotic intelligence.

One of the principal research directions highlighted in the interviews is grasp identification. Deep learning has made strides here as well: the Dex-Net 2.0 model uses depth images to identify grasp points, and has outperformed baselines based on registration or image-based heuristics and demonstrated effectiveness on a series of unseen objects (Mahler *et al.*, 2017). Several other methods have yielded similar successes: Bousmalis *et al.* (2018) used a simulation to real-world pipeline consisting only of RGB cameras to successfully perform grasping, while Zeng *et al.* (2018) described a method involving fully convolutional networks to identify grasp points irrespective of object type. Others have involved the integration of vision and grasping pipelines via deep-learning approaches (Fryman and Matthias, 2012; Levine *et al.*, 2016; Quillen *et al.*, 2018).

Dexterous in-hand manipulation represents another promising area for research. A number of existing approaches in literature used tactile sensing for in-hand manipulation(Lee, 2000; Tian *et al.*, 2019) (Yousef *et al.* (2011) provided a review of such techniques). In the context of bin picking, Correll *et al.* (2016) reported observations and challenges from the Amazon Picking Challenge in 2015, which required teams to integrate solutions related to object perception, motion planning, grasp planning, and task planning, and emphasized the importance of reliability. Fujita *et al.* (2019) provided further overview of important technologies for bin picking, including gripper design and grasp planning.

However, it appears that such advances have yet to be integrated into industrial environments. The requirements for extreme robustness and reliability represent particular hurdles here; with this in mind, we note the following: the majority of robots in industrial environments operate with a limited number of possible workpieces and perform repetitive, scripted motions. Manufacturers hope to reduce their dependency upon both custom grippers and the infrastructure required to orient and position parts for robotic collection. One valuable direction for further research could be a method for robustly identifying, manipulating, and grasping a known object that includes size and pose variations, with the final grasp orientating the object in a specific configuration.

## 7.3 Collaborative "Safe" Robots

The definition of "collaborative roboticsâĂİ appears to differ substantially between industry and academia. Industrial systems considered "collaborativeâĂİ by many interviewees were primarily robot workzones that humans could safely enter. While this subject has already received significant attention (see Lasota *et al.* (2017) for an extensive review), collaborative industrial robots could benefit from the idea of *anticipation* (Hoffman and Breazeal, 2007): the integration of prediction of a human's action or goal into a robot's action or path planning. Such methods have previously been used to dynamically avoid a human operator (Mainprice and Berenson, 2013) or plan a robot's actions in response to a perceived goal (Hoffman and Breazeal, 2007; Lasota *et al.*, 2014; Freedman and Zilberstein, 2017). The idea of such methods is to either enable new capabilities by allowing dynamic collaboration between a human operator and the robot, or to further ensure the safety of a human associate by reducing potential proximity to the robot while also seeking to improve the performance of the human-machine system.

The lack of use of such systems within industrial environments is likely due to several factors: first, these systems require monitoring by a human associate, a consideration that may have social implications and severe instrumentation and perception requirements. Second, collaborative workcells, by definition, differ greatly from typical robotics applications, and traditional cost analyses (such as labor displacement) may not adequately reflect the benefit of such cells. Finally, safety remains a concern, and risk assessment of a dynamic workcell may be difficult to perform. To this end, we suggest several directions for future research: to demonstrate the advantages and benefits of collaborative workcells, highlight classes of human-led industrial work that could benefit from robotic assistance, and the development of technologies for the robust perception and tracking of a human associate within a workcell, which could improve integration of existing methods for dynamic collaboration.

Safety standards represent another interesting direction for research. While standards do exist (such as ISO15066 (ISO, 2016)), much of the interview feedback centered around performance restrictions imposed by those standards. This involves a nuanced discussion: without the restrictions, harm could occur to humans operating within a robot's environment; but with restrictions, the robot is comparatively less effective at performing its tasks. It is important to note that, irrespective of the safety systems integrated with the robot (such as the ability to detect a human before contact is made), the velocity requirements of the standards still apply. An interesting research topic may be qualifying the safety of a robotic system in the context of additional safety mechanisms (such as those discussed in Section 3) with the hope of producing provable safety criteria that do not otherwise limit robots' capabilities.

Finally, several interview subjects ([RM1], [PM1] and [RM2]) noted that system safety is not viewed only with regard to the robot, but to the system as a whole. Much of the discussion about safety for a human/robot system operating in a common workzone is focused on ensure that the given task is safe for the human to interact with: for example, a robot moving objects with sharp edges, or performing tasks (such as welding) that would be hazardous to a human in close proximity are not open to the application of collaborative systems. Improving the safety of hazardous tasks that a robot may perform to ensure that a human could share a workspace with the robot may also be a valuable research topic. Several approaches have attempted to ensure the safety of a human operator during such operations: Peternel *et al.* (2018) developed a method based on dynamic movement primitives to account for operator fatigue, while Lamon *et al.* (2020) presented a framework to enable robotic assistance during palletizing processes, and Zanchettin *et al.* (2019) developed an optimal collision-avoidance scheme using models of human motion.

## 7.4   Autonomous Guided Vehicles

As discussed earlier, AGVs were a common topic during the interviews; however, their overall value was disputed, and technological challenges remain despite their increasing popularity. AGVs have been discussed frequently within academia as well, particularly within the simultaneous localization and mapping (SLAM) and autonomous navigation communities.

SLAM technologies enable a robot to generate a precise map of its environment. The technology is based on estimation theory, originally using Kalman filters (Dissanayake *et al.*, 2001), and later more optimal factor graph-based solutions (Kaess *et al.*, 2008). While a solution to the SLAM problem can be computed, challenges remain; data association (the act of associating detected features or landmarks with previous measurements) has proven particularly difficult, prompting the development of robust optimization methods (Sünderhauf and Protzel, 2012), or probabilistic association (Bowman *et al.*, 2017). This challenge can be overcome with effective and robust navigation techniques, enabling deployment of fleets of autonomous robots within a factory environment. However, the existing navigation capability of a service robot utilizing this technology does not appear sufficiently precise for mobile manipulation. Several interview subjects mentioned a desire for manipulators to be integrated onto mobile bases; however, this itself poses significant challenges: in particular, it appears to highlight a need for intelligent gripping to account for position estimation errors, or precise localization within the range of the repeatability of the robot arm (typically sub-millimeter for a standard industrial arm).

The precise benefits of industrial application of mobile robots differed significantly between interviews, and their implementation on factory floors appears to be largely experimental, with the value proposition primarily prospective. A valuable direction for future research may lie in defining metrics for assessing the performance of a fleet of mobile robots, as well as defining tasks for which their integration could yield tangible benefits.

## 7.5 Interfaces and Programming

Easy-to-use robot interfaces and simpler robot programming emerged as a key research direction for most companies. [PM1] discussed learning from demonstration, kinesthetic teaching, and semantic programming interfaces as potential research directions in particular; we view these and others as promising directions for the research community to explore.

Prior research into human-robot interaction (HRI) has sought to enable easy-to-program robots that do not require significant programming expertise for use. Research into learning from demonstration (LfD) (also referred to as programming by demonstration [PbD]) reviewed by Argall *et al.* (2009), Billard *et al.* (2008), and more recently by Zhu and Hu (2018) (who highlighted the assembly context in particular) has addressed this challenge. In LfD, users demonstrate a sequence of robot actions through a set of concrete examples, with the robot then using these examples to learn and generalize its behavior to new examples. A number of modalities have previously been applied to providing demonstrations, including kinesthetic teaching, natural language, and visual programming interfaces, among others (Ajaykumar and Huang, 2020). Kinesthetic teaching is a widely used approach that enables users to program a robot by physically guiding it through a task; however, a recent study found that human users can experience fatigue and a lack of adequate mental models of the robot's learned behavior with such a method (Ajaykumar and Huang, 2020).

One recent technique aims to build upon current kinesthetic teaching and LfD techniques by integrating LfD with task-level programming, which provides semantic information associated with robot tasks to enhance users' understanding of robot behavior (Steinmetz *et al.*, 2019). Other methods in explainable AI and related fields can also help align user mental models with the robot's behavior, particularly with regard to enabling predictable robot actions, as discussed in work by (Dragan *et al.*, 2013). Approaches such as active learning require smaller amounts of data and improve learning rates by asking users to provide specific types of demonstrations or information (Cakmak and Thomaz, 2012). Such approaches are helpful for enabling learning and generalization with scarce available data, as is common in applications where humans provide demonstrations; they can also help reduce the burden upon and fatigue of human demonstrators. However, effective learning from limited data remains an open challenge for LfD and related applications.

In addition to the approach introduced by Steinmetz *et al.* (2019), other recent work has similarly focused on semantic-level robot programming, which can allow workers to more readily "program the task and not the robot" - a goal [PM1] and [PM2] considered important. Perzylo *et al.* (2015) and Perzylo *et al.* (2016) proposed programming paradigms that enable users to use semantic descriptions of work processes, workpieces, and workcells along with a graphical programming interface to program the robot. Such semantic programming interfaces can hasten robot programming by lowering the barrier to entry (in terms of necessary skills and training) to control such systems. To this end, valuable future research directions include exploring semantic programming interfaces at greater length, as well as LfD approaches that support users' mental models and limited data applications. Shortening the simulation-to-programming pipeline represents another valuable avenue of research, and could be enabled through higher fidelity and more integrated simulations.

## 7.6 Simulation

As discussed in Section 6.3, enhanced simulations can reduce the time necessary to program and integrate robots into workcells, as well as contribute to production optimization and improve the use of collected data about tools and processes. The research community can address challenges related to the use, applicability, and capabilities of simulation tools; for example, some recent work proposed a probabilistic inference technique to estimate uncertain model parameters for simulations (Ramos *et al.*, 2019). Such approaches could contribute to higher-fidelity simulation of robots for manufacturing lines, especially when precise modeling is difficult.

While high-fidelity simulations and enhanced digital twins could improve the accuracy of production estimates in individual processes for manufacturing lines, integrating simulations at multiple levels of abstraction (including the workcell, manufacturing line, and plant levels, as discussed by [PM2]) also represents a challenge. To this end, Mourtzis *et al.* (2014) and Mourtzis (2019) provided surveys of technologies and research directions related to simulating manufacturing systems (and the integration of simulations) at multiple levels. Such simulations could help reduce the burden of robotics integration (as discussed in Section 5) and provide better estimates of entire life-cycle costs for robotics technologies (Mourtzis *et al.*, 2014). The best way in which to use collected data to improve simulations in both a computationally tractable and useful manner remains an open question

Virtual reality (VR) and augmented reality (AR) systems, among other types of simulations, demonstrate promise in terms of reducing integration times and shortening the simulation-to-programming pipeline by enabling line workers and others to quickly prototype potential solutions without needing to program robots at a low level. Burghardt *et al.* (2020) introduced an integrated VR and digital twin system for programming industrial robots, while results from work by Kapinus *et al.* (2020) and Gadre *et al.* (2019) suggested that mixed- and augmented-reality interfaces using head-mounted displays reduce integration time and user workload and improve usability compared with traditional programming interfaces. However, implementation challenges remain for both VR and AR systems, including extensive infrastructure requirements and human factors-related issues (such as motion sickness).

## 7.7   Worker-Centered Design

Worker-centered design is another key research topic that could both enhance worker experience when interacting with robots and improve productivity and efficiency of workflows. Worker-centered design focuses on the needs and preferences of workers (often elicited through conversations and interviews) in the design and development of workcells for manufacturing lines. Existing interview- and survey question-based research has explored workers' perspectives of robots in a production environment over time (including pre-introduction, familiarization, and experienced consequences) (Wurhofer *et al.*, 2015), workers' experiences related to ergonomics with new technologies (Colim *et al.*, 2020), and the social impacts of introducing robots as co-workers into manufacturing settings (Sauppé and Mutlu, 2015; Meneweger *et al.*, 2015), among others. Moniz and Krings (2016) identified additional questions that are important to explore in the context of worker-centered design, such as how intuitive systems are and whether workers trust the robots.

While our interviews suggest a number of companies believe involving workers in the design and integration processes for new technologies can contribute to worker performance and ultimately improve manufacturing line productivity, little research has been performed to concretely demonstrate this link. Huber and Weiss (2017) compared workers' performance expectancy for two robots with different interfaces, while other work has further explored human factors and performance-related considerations for manufacturing systems (Kadir *et al.*, 2019; Pacaux-Lemoine *et al.*, 2017). Overall, empirically demonstrating the performance- and productivity-related improvements to be gained from increased worker input into the design process represents another viable direction for future research.

In addition to interview-based approaches to worker-centered design, technical solutions can also be employed to enhance worker-robot interactions. For example, easier-to-use and easier-to-program interfaces (as discussed in Section 6.1) can reduce training requirements and allow workers to more quickly use these systems for their purposes. Beyond this, systems that can infer worker preferences and work styles, as discussed by Iqbal *et al.* (2019)

and Fourie (2019), can make robots more adaptable to their human partners. Since human line workers possess deep domain knowledge and often understand how manufacturing processes can be improved (as indicated in several of our interviews), it is important that any future technologies can capture and integrate this knowledge and leverage human strengths and creativity. We recommend further research into technical approaches that enable robots to adapt to humans' preferences and individual needs.

## 7.8 Cloud Systems

Ideas around Industry 4.0, the Internet of Things (IoT), and cloud systems in general were a common topic during the interviews. While these ideas appear attractive from a general perspective, our interview subjects largely expressed uncertainty about the benefits of investing in such technologies. Many cited the cost of wide-scale sensor integration, while others acknowledged data security and ownership concerns. The predominant limitation, however, was that the direct benefit to collecting data remains unclear.

IoT and cloud systems have received significant interest from the academic community. Recent work has discussed the architectural challenges of processing large amounts of data in real time (Cheng *et al.*, 2018), while findings from a recent survey suggest that communication protocols used in IoT and cloud computing are fragmented, and that multiple competing solutions exist (Dizdarević *et al.*, 2019). Many enterprise-level systems exist for accessing data from industrial automation systems, the most popular of which appears to be OPC UA, a unified architecture for accessing data and events developed by the OPC Foundation (Bangemann *et al.*, 2014).

However, feedback from the industrial community highlights several interesting research directions related to use cases for the collected data. This includes fault detection and isolation, predictive maintenance, and system performance reporting. While [RM1], [PM2], [RI2], and [PM1] all discussed the uses of such systems within their own industrial context, they also highlighted the difficulty of utilizing these systems for wide-scale performance monitoring and predictive maintenance. We believe methods for automatic performance and maintenance monitoring techniques with minimal configuration requirements represent valuable potential avenues for future research. In addition, [RI1] discussed the "island problem" in detail, where systems are often discon-nected, with no easy path for integration into a unified framework; addressing this challenge may yield novel research directions for interconnected systems. Finally, [RI2], and [RM2] also noted resistance within the industry to the instrumentation of production lines, citing unclear benefits and data security concerns. Addressing security issues could pose interesting challenges for the academic community, while exploring novel use cases for the collection of large amounts of industrial data could also be of extraordinary benefit.

# 8

---

# Related Work

---

Prior work in the robotics research community has identified challenges related to integrating robotics into industrial manufacturing (Hentout *et al.*, 2018; Hentout *et al.*, 2019). These works have primarily focused on human-robot interaction: Hentout *et al.* (2019) in particular provided a comprehensive literature review of research relevant to industrial robotics. While these works enumerate some of the challenges and relevant research problems for manufacturing robotics, we view the perspectives of key players in the robotics ecosystem beyond the research community as critical to understanding the drivers behind important areas of focus in robotics technology development and which of these represent primary focuses in industry.

Further, we note some overlap between the challenges identified in Hentout *et al.* (2018) and Hentout *et al.* (2019) and those discussed in our interviews, particularly in terms of important future directions for robotics technology development. However, these prior works lack a broader ecosystem perspective related to integration of robotics into manufacturing lines and the standardization of many aspects of robotics architectures. Elprama *et al.* (2016), Elprama *et al.* (2017), Sauppé and Mutlu (2015), Welfare *et al.* (2019), and Wurhofer *et al.* (2018) interviewed manufacturing employees who work closely with new robotics technologies, and included line workers' perspectives on the introduction of these technologies in their analyses. Although we consider the line worker's viewpoint to be valuable, we aim to represent a broader perspective informed by decision makers at multiple levels. While some works have discussed safety standards (Eder *et al.*, 2014; Fryman and Matthias, 2012; Michalos *et al.*, 2015) or communication and infrastructure standards (Chen, 2017; Monostori *et al.*, 2016), a more comprehensive discussion of standardization illustrated by key players in the robotics ecosystem (as presented in Section 5) is absent from current literature, to our knowledge. Beyond this, while we confirm some of the important technological directions for robotics development and challenges posed by Hentout *et al.* (2019) through our interviews, our subjects identified technological bottlenecks based on real-world application of these technologies that go beyond what the research community has identified in some cases (for example, the importance of industrial-level robustness and reliability of robotics solutions for manufacturing).

# 9

# Conclusion

In this paper, we detailed results from our interviews of various actors in the robotics ecosystem - including robot manufacturers and integrators, research institutions, and product manufacturers. We identified primary themes that emerged during these interviews with regard to current and emerging technologies, company processes related to adoption of new robotics technologies, challenges related to this adoption of new robotics technologies, and next company directions in terms of robotics technology development.

While standard industrial robotics design has undergone few changes in recent years, development thus far appears to have focused on supplemental systems (e.g., safety systems, compliant robotic arms, and improved interfaces) primarily intended to reduce integration costs or improve safety. Lightweight robotics, a low-cost and low-payload alternative to the standard industrial robot, has been demonstrated as a viable, safe alternative for some processes, enabling simpler integration and new workflows. Autonomous guided vehicles (AGVs) have also been beneficial for industrial logistics, while integration of IoT technologies - and cloud systems in particular - has been hampered by security concerns and an unclear value proposition.

Several key themes were clear throughout the interviews: first, decisions about whether to apply automation are primarily based on total cost or return on investment over the lifetime of the technology, although considerations such as ergonomics, an aging workforce, and quality control were also important factors. Second, worker perspectives are critical to automation, as it is often the voice of the worker that drives innovation. Finally, focus appears to be shifting toward increased flexibility, achieved either by empowering workers to work directly with automation or by improving the way in which engineers interface with robotic workcells (explicitly defined as "programming the task and not the robot" by a number of companies throughout the interviews).

Robotic automation also faces several challenges. Long and expensive integration work, coupled with a general scarcity of integrators, delays product development and reduces flexibility. Another significant hurdle appears to be a lack of standardization in general, limiting companies' ability to reuse robotic systems, inhibiting modularity, and increasing integration time. A lack of flexibility in production is also a significant issue, with companies unable to easily reuse robotics or adapt robotic solutions to new workpieces. Technological bottlenecks related to sensing, perception, and gripping are also apparent, largely driven by a need for extreme robustness and reliability that even recent innovations (such as deep learning) have not yet addressed.

Finally, we outlined recommendations for the research community related to processes, challenges, and the next areas of focus for robotics development under consideration by companies within the manufacturing ecosystem. As these recommendations are the product of feedback from the industrial robotics community, they represent high-value directions for future academic focus.

# Acknowledgments

This work was supported by the MIT Work of the Future Task Force. The authors would particularly like to thank the Work of the Future backbone group for conversations about this work as well as Suzanne Berger, Liz Reynolds, David Mindell, Anna Waldman-Brown, and Dan Traficonte for all of their insights.